\documentclass{article}

\usepackage{microtype}
\usepackage{graphicx}
\usepackage{subcaption}
\usepackage{booktabs} 

\usepackage{hyperref}
\usepackage{booktabs}
\usepackage{multirow}
\usepackage{tikz}
\usepackage{pgfplots} 
\pgfplotsset{compat=1.13}
\newcommand*\circled[1]{\tikz[baseline=(char.base)]{
            \node[shape=circle,draw,inner sep=0.5pt] (char) {#1};}}

\definecolor{bblue}{HTML}{4F81BD}
\definecolor{rred}{HTML}{C0504D}
\definecolor{ggreen}{HTML}{9BBB59}
\definecolor{ppurple}{HTML}{9F4C7C}

\newcommand{\rowfont}[1]{
   \gdef\rowfonttype{#1}#1%
   }

\newif\ifarxiv

\arxivtrue 

\newcommand{\arxivornot}[2]{%
  \ifarxiv
  #1 
  \else
  #2 
  \fi
}

\arxivornot{
    \usepackage[accepted]{icml2024}
}
{
    \usepackage{icml2024}
}

\usepackage{amsmath}
\usepackage{amssymb}
\usepackage{mathtools}
\usepackage{amsthm}
\usepackage{thmtools, thm-restate}
\usepackage{bbm}
\usepackage{nccmath}
\usepackage{tabularx}

\usepackage[capitalize,noabbrev]{cleveref}

\theoremstyle{plain}

\theoremstyle{definition}

\theoremstyle{remark}

\usepackage[textsize=tiny]{todonotes}

\newcommand{\V}{\mathcal{V}}
\newcommand{\Y}{\mathcal{Y}}
\newcommand{\X}{\mathcal{X}}
\newcommand{\y}{\boldsymbol{y}}
\newcommand{\ysamp}[1]{\boldsymbol{y}_{#1}}

\newcommand{\rsamp}[1]{r_{#1}}
\newcommand{\x}{\boldsymbol{x}}

\newcommand{\betar}{{\beta_r}}
\newcommand{\betap}{{\beta_\pi}}
\newcommand{\rhuman}{r_{{h}}}

\newcommand{\pisft}{{\pi_{\textup{sft}}}}

\newcommand{\pifkl}{\pi_{\theta_{\textup{FKL}}}}
\newcommand{\pirkl}{\pi_{\theta_{\textup{RKL}}}}
\newcommand{\piinit}{\pi_{\textup{init}}}

\newcommand{\Datasft}{\mathcal{D}^{\textup{sft}}}
\newcommand{\Datapref}{\mathcal{D}^{\textup{pref}}}
\newcommand{\Datarw}{\mathcal{D}^{\textup{rw}}}
\newcommand{\Data}{\mathcal{D}}

\newcommand{\Lreward}{\mathcal{L}_{r}}
\newcommand{\Jlhf}{\mathcal{J}_{\textup{lhf}}}
\newcommand{\Ldpo}{\mathcal{L}_{\textup{dpo}}}
\newcommand{\Lexact}{\mathcal{L}_{\textup{exo}}}

\newcommand{\Lexactpref}{\mathcal{L}_{\textup{exo-pref}}}
\newcommand{\Ldpogeneral}{\mathcal{L}_{\textup{dpo-rw}}}

\newcommand{\KLDiv}{\mathbb{D}_{\textup{KL}}}

\newcommand{\EXOpref}{{$\textup{EXO}_{\textup{pref}}$}}
\newcommand{\DPOpref}{{$\textup{DPO}_{\textup{pref}}$}}
\newcommand{\EXOrw}{{$\textup{EXO}_{\textup{rw}}$}}
\newcommand{\DPOrw}{{$\textup{DPO}_{\textup{rw}}$}}

\icmltitlerunning{Towards Efficient Exact Optimization of Language Model Alignment}

\begin{document}

\twocolumn[
\icmltitle{Towards Efficient Exact Optimization of Language Model Alignment}



\icmlsetsymbol{equal}{*}

\begin{icmlauthorlist}
\icmlauthor{Haozhe Ji}{coai}
\icmlauthor{Cheng Lu}{tsail}
\icmlauthor{Yilin Niu}{zhipu}
\icmlauthor{Pei Ke}{coai}
\icmlauthor{Hongning Wang}{coai}
\icmlauthor{Jun Zhu}{tsail}
\icmlauthor{Jie Tang}{keg}
\icmlauthor{Minlie Huang}{coai}
\end{icmlauthorlist}

\icmlaffiliation{zhipu}{Zhipu AI}
\icmlaffiliation{coai}{The Conversational AI (CoAI) Group, Tsinghua University}
\icmlaffiliation{tsail}{The Tsinghua Statistical Artificial Intelligence \& Learning (TSAIL) Group, Tsinghua University}
\icmlaffiliation{keg}{The Knowledge Engineering Group (KEG), Tsinghua University}

\icmlcorrespondingauthor{Haozhe Ji}{jihaozhe@gmail.com}
\icmlcorrespondingauthor{Minlie Huang}{aihuang@mail.tsinghua.edu.cn}

\icmlkeywords{Machine Learning, ICML}

\vskip 0.3in
]



\printAffiliationsAndNotice{}  

\begin{abstract}
The alignment of language models with human preferences is vital for their application in real-world tasks. The problem is formulated as optimizing the model's policy to maximize the expected reward that reflects human preferences with minimal deviation from the initial policy. While considered as a straightforward solution, reinforcement learning (RL) suffers from high variance in policy updates, which impedes efficient policy improvement. 
Recently, direct preference optimization (DPO) was proposed to directly optimize the policy from preference data. 
However, we show that DPO derived based on the optimal solution of the problem 
leads to a compromised mean-seeking approximation of the optimal solution in practice. 
In this paper, we propose \textit{efficient exact optimization} (EXO) of the alignment objective. 
EXO is guaranteed to optimize in the same direction as RL algorithms asymptotically for arbitrary policy parametrization. This leads to the same mode-seeking solution, while enables efficient optimization by circumventing the complexities of RL. 
We also compare our method to DPO with both theoretical and empirical analyses, and further demonstrate the advantages of our method over existing approaches on realistic human preference data. 
\arxivornot{
Code is available at \url{https://github.com/haozheji/exact-optimization}.
}{}
\end{abstract}

\section{Introduction}

Despite the proficiency of large language models, e.g., GPT-3~\citep{gpt-3} in complex tasks under minimal supervision, 
they are still prone to produce harmful~\citep{hh}, biased~\citep{stochastic-parrots}, and unfaithful~\citep{ji2023survey} responses due to the heterogeneous sources of their pre-training corpora. 
Ensuring the large language models to generate desired responses that are in line with humans' ethical standards and quality preferences is crucial for the development of reliable AI systems.

The problem, well known as language model (LM) 
alignment with human preferences~\citep{instruct_gpt}, is generally formulated as optimizing the LM policy $\pi_\theta$ to maximize the expected reward, which reflects human preferences regarding the completion $\y$ for a given prompt $\x$. The practical recipe is to train a reward model $r_\phi$ to predict the human-chosen response from a set of responses generated by an initial LM policy $\piinit$. Yet, the challenge of acquiring substantial high-quality preference data often impedes accurate estimation of the ideal reward model. 
Consequently, this empirically learned reward model may lead to misspecified behaviors, particularly under the distributional shift 
between its training data 
and the data generated by $\pi_\theta$~\citep{overoptimization}. Therefore, the final objective of alignment additionally involves minimizing the reverse Kullback-Leibler (KL) divergence of $\pi_\theta$ from its initial distribution $\piinit$ with an intensity $\beta$, besides maximizing the expected reward:
\begin{equation}
    \label{equ:intro_rlhf}
    \max_{\pi_\theta}
    \mathbb{E}_{\x\sim\Data,\y\sim\pi_\theta(\y|\x)}[r_\phi(\x, \y)] - \beta \KLDiv(\pi_\theta \| \piinit).
\end{equation}
Due to the discrete nature of content generation from an LM (i.e., sampling $\y$ from $\pi_\theta(\cdot|\x)$), the most widely adopted approach to optimize Eq. \eqref{equ:intro_rlhf} is reinforcement learning (RL)~\citep{ziegler_rlhf,summarize_rlhf,instruct_gpt}. 
Albeit well studied even before the era of large language models, RL solutions are notorious for their poor stability due to the high variance in estimating the policy gradients or value functions, which potentially worsens sample complexity and thus compromises efficient convergence \cite{papini2018stochastic,anschel2017averaged}. 

As a remedy, direct preference optimization (DPO) was recently proposed to replace the RL solutions~\citep{rafailov2023direct}. Specifically, DPO defines a pair-wise preference loss on the estimated policy $\pi_\theta$ by leveraging the following policy-reward mapping in the optimal solution to Eq. \eqref{equ:intro_rlhf}:
\begin{equation}
    \label{equ:intro_optimal}
    \pi^{*}_{\beta}(\y|\x)\propto \piinit(\y|\x) e^{\frac{1}{\beta}r_\phi(\x,\y)}
\end{equation}
However, this optimal policy is obtained analytically while not considering the practical parametrization of $\pi_\theta$.
When $\pi_\theta$ is mis-specified and cannot perfectly capture the target $\pi_\beta^*$, our analysis demonstrates that DPO 
leads to a compromised approximation of $\pi_\beta^*$, which only covers the support of $\pi_\beta^*$ while failing to capture its modes.

In this paper, we present an underexplored perspective of the alignment objective in Eq. \eqref{equ:intro_rlhf}: We prove that Eq. \eqref{equ:intro_rlhf} is equivalent to \textit{probability matching} between the parametrized policy $\pi_\theta$ and the optimal policy $\pi_{\beta}^*$ defined in Eq. \eqref{equ:intro_optimal} measured by the reverse KL divergence $\KLDiv(\pi_\theta\|\pi_{\beta}^*)$.

Based on the established equivalence, we propose \textit{efficient exact optimization (EXO)} of the KL-regularized reward maximization objective.  
Specifically, we rigorously prove that irrespective of the policy's parametrization, EXO is guaranteed to improve $\pi_\theta$ in the same direction as the RL solutions asymptotically. In practice, we demonstrate that EXO facilitates efficient optimization towards this alignment objective with finite samples while bypassing the complexities of RL. 



\begin{figure}
    \centering
    \includegraphics[width=0.985\columnwidth]{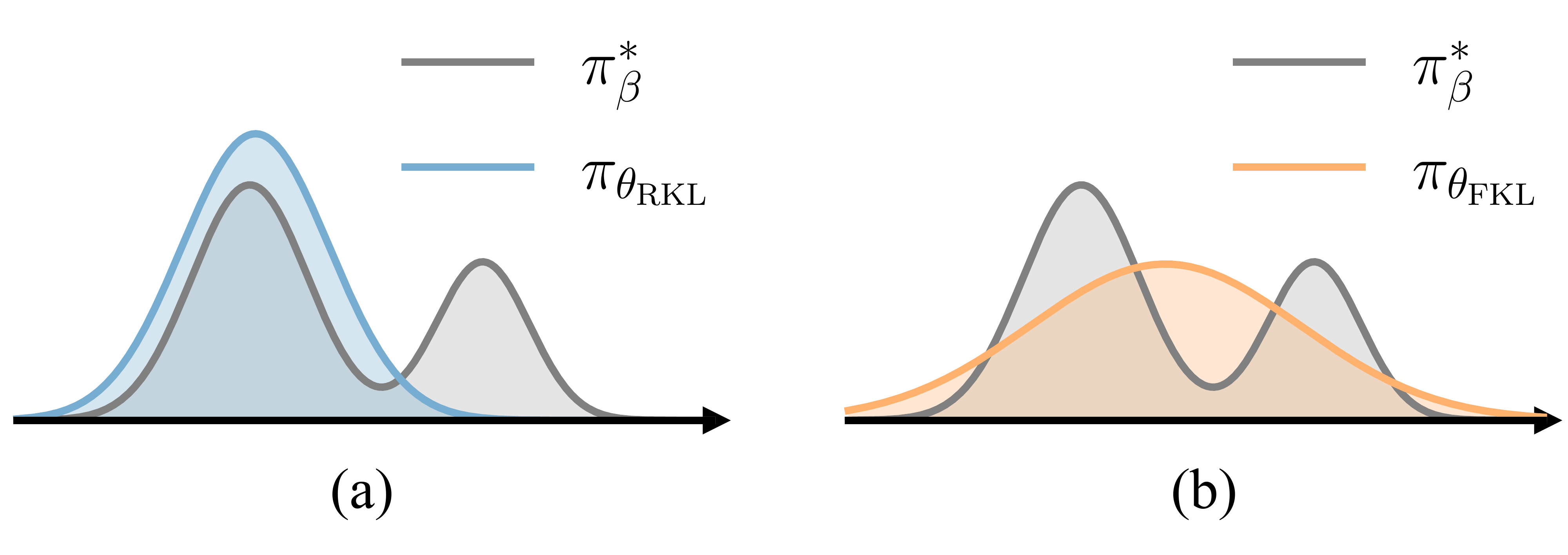}
    \caption{
    Illustration of different characteristics of (a) $\pirkl$ by minimizing the reverse KL (by EXO) and (b) $\pifkl$ by minimizing the forward KL (by DPO).
    }
    \label{fig:fwkl_revkl}
\end{figure}

Under this probability matching perspective, we demonstrate that DPO actually 
corresponds to minimizing the forward KL divegence $\KLDiv(\pi_{\beta}^*\|\pi_\theta)$. 
Though minimizing both the forward and reverse KL divergences lead to the same analytic solution, it is not necessarily achievable when taking into account the expressivity gap between the model families of $\pi_{\beta}^*$ and $\pi_\theta$~\citep{Lin2021LimitationsOA}. 
Under this realistic constraint, minimizing these two divergences converge to parametrized policies with different behaviors~\citep{bishop2006pattern}. 
As illustrated in Figure \ref{fig:fwkl_revkl}, minimizing the reverse KL fosters a \textit{mode-seeking} policy $\pirkl$ that concentrates to the principal modes of $\pi_\beta^*$~\citep{chan2022greedification}, while minimizing the forward KL results in a \textit{mean-seeking} policy $\pifkl$ that places large mass to the mean of different modes in $\pi_{\beta}^*$, which does not necessitate high probabilities under $\pi_{\beta}^*$. 
In the inference stage, $\pirkl$ is preferrably better than $\pifkl$ by capturing the main characteristics of $\pi_{\beta}^*$~\citep{daemon}.

We conduct a series of experiments to verify the effectiveness and scalability of EXO. We first systematically evaluate the efficiency of different approaches in trading off maximizing the oracle reward and minimizing the KL divergence during optimization of the alignment objective. 
Then, we conduct evaluations on the effectiveness of learning from real human preferences in various alignment benchmarks, involving summarization, dialogue generation, and instruction following tasks.
Comprehensive empirical analyses substantiate our theoretical findings and demonstrate the advantageous performance of EXO over DPO and PPO.



Finally, we summarize our contributions in this paper:
\begin{itemize}
    \item We reveal the underexplored equivalence between KL-regularizd reward maximization  and minimizing the reverse KL divergence against the optimal policy for the language model alignment problem.
    \item We propose EXO, an algorithm towards \textit{efficient exact} optimization of the KL-regularized reward maximization objective for alignment. Both theoretical and empricial results confirm its effectiveness.
    \item We show that DPO corresponds to minimizing the forward KL divergence, which is less effective in capturing the essential characteristics of the optimal policy.
\end{itemize}


\section{Preliminaries}

We first formally review the formulation and objective of the alignment problem. Then we review existing approaches that solve this problem via reinforcement learning and 
direct preference optimization, respectively. 

\subsection{Aligning Language Models with Human Preferences}

Given a vocabulary $\V$, a language model defines a probability distribution $\pi(\x)=\prod_{t=1}^n \pi(x_t|x_1,\cdots,x_{t-1})$ over a sequence of tokens $\x=(x_1,\cdots,x_n)$. 
We apply $\pi$ to a conditional generation task of interest with input space $\X=\V^m$ and output space $\Y=\V^n$ modeled by $\pi(\y|\x)=\pi(\x,\y)/\pi(\x)$. 

The alignment procedure typically starts from supervised fine-tuning (SFT) the language model on a high-quality dataset $\Datasft$ via maximum likelihood estimation, which obtains the SFT policy $\pisft$.

Then a preference dataset $\Datapref$ is collected by asking humans to select a better response from $(\ysamp{1},\ysamp{2})\sim \pisft(\y|\x)$ given a prompt $\x$ from the same domain of $\Datasft$. Let $\y_w$ and $\y_l$ be the chosen and rejected responses among $(\ysamp{1},\ysamp{2})$ respectively according to human preferences. 

A reward model $r_\phi: \X\times \Y\rightarrow \mathbb{R}$ is usually learned on $\Datapref$ to act as a surrogate to expensive human labeling. 
The reward model is trained to prioritize $\y_w$ over $\y_l$ by minimizing the following pair-wise preference loss:
\begin{equation}
    \label{equ:reward_loss}
    \Lreward(r_\phi) = \mathbb{E}_{(\x,\y_w,\y_l)\sim \Datapref}\bigg[-\log \frac{e^{r_\phi(\x,\y_w)}}{e^{r_\phi(\x,\y_w)} + e^{r_\phi(\x,\y_l)}}\bigg].\nonumber
\end{equation}
Finally, a policy $\pi_\theta$ is learned to maximize the following alignment objective~\citep{ziegler_rlhf}:
\begin{align}
    \Jlhf^\beta(\pi_\theta) &= \mathbb{E}_{\x\sim \Datapref}\Big( \mathbb{E}_{\pi_\theta(\y|\x)}[r_\phi(\x,\y)] \nonumber\\
    &\quad \quad - \beta \KLDiv[\pi_\theta(\y|\x)\|\pisft(\y|\x)]\Big).
    \label{equ:lhf_objective}
\end{align}
Given a prompt $\x$ sampled from the dataset $\Datapref$, the objective seeks to find the $\pi_\theta$ that maximizes the expected reward  while minimizes its reverse KL divergence against the SFT policy $\pisft$ governed by the coefficient $\beta> 0$. The KL penalty keeps $\pi_\theta$ from moving too far from $\pisft$ to avoid over optimization of the reward model.

The analytic solution 
that maximizes $\Jlhf^\beta(\pi_\theta)$ takes the form of an energy-based model (EBM):
\begin{equation}
    \label{equ:optimal}
    \pi_{\beta}^*(\y|\x) = 
    \pisft(\y|\x)
    \frac{
        e^{\frac{1}{\beta}r_\phi(\x,\y)}
    }{Z_\beta(\x)},
\end{equation}
where $Z_\beta(\x)=\sum_{\y'\in \mathcal{Y}}\pisft(\y'|\x)e^{\frac{1}{\beta}r_\phi(\x,\y')}$ is the partition function. In Eq \eqref{equ:optimal}, the coefficient $\beta$ can be considered as the temperature for controlling the strength of the reward model signal when sampling from $\pi_{\beta}^*(\y|\x)$. 

\subsection{RL Fine-Tuning}

Due to the discrete nature of language generation, the objective in Eq. \eqref{equ:lhf_objective} is not differentiable with respect to $\pi_\theta$, which prohibits supervised training. One standard approach is to use RL algorithms to optimize this objective. \citet{ziegler_rlhf} proposed to search for $\pi_\theta$ that maximizes a KL-regularized reward $r_\phi(\x,\y) - \beta\log\frac{\pi_\theta(\y|\x)}{\pisft(\y|\x)}$.  
This can be achieved by policy gradient methods, such as Proximal Policy Optimization (PPO)~\citep{schulman2017proximal}.

\subsection{Direct Preference Optimization}
\label{sec:DPO}
To optimize $\pi_\theta$ directly using the preference data, \citet{rafailov2023direct} rearranged Eq. \eqref{equ:optimal} to express the reward function by the optimal policy $\pi_{\beta}^*$, 
\begin{equation}
\label{equ:dpo-reward}
    r_\phi(\x,\y) = \beta \log \frac{\pi_{\beta}^*(\y|\x)}{\pisft(\y|\x)} + \beta\log Z_\beta(\x).
\end{equation}
Then they proposed to directly optimize the policy $\pi_\theta$ by replacing $\pi_\beta^*$ with $\pi_\theta$ and substituting the corresponding reward function into a pair-wise preference loss:
\begin{align}
    \label{equ:dpo}
    &\Ldpo(\pi_\theta) =\mathbb{E}_{(\x,\y_w,\y_l)\sim \Datapref}\bigg[\nonumber\\
    &\quad -\log\sigma\Big(
    \beta\log \frac{\pi_\theta(\y_w|\x)}{\pisft(\y_w|\x)}
    - \beta\log \frac{\pi_\theta(\y_l|\x)}{\pisft(\y_l|\x)}
    \Big) 
    \bigg].
\end{align}

\section{Methodology}
\label{sec:method}


In this section, we start with a generalized alignment objective and present its equivalent form under the perspective of probability matching. Then we formally derive 
\textit{efficient exact optimization} (EXO) of the generalized alignment objective while bypassing the necessity of employing any RL algorithms. Furthermore, we compare against DPO 
and demonstrate the advantage of our approach in terms of the distributional characteristics of the learned policy. All the proofs are provided in Appendix \ref{appendix:proofs}.




\subsection{From the Generalized Alignment Objective to Probability Matching}
\label{sec:equivalence}

We first introduce a generalized alignment objective that distributes the intensity of the KL regularization regarding the SFT policy to both the parametrized policy $\pi_\theta$ and the reward model $r_\phi$, which intuitively connects the regularization setting of DPO (Eq. \eqref{equ:dpo}) that only regularizes $\pi_\theta$ and PPO (Eq. \eqref{equ:lhf_objective}) that only regularizes $r_\phi$. 
In the following theorem, we present the formal definition and the property of the generalized alignment objective.
\begin{restatable}{theorem}{GeneralizedTheorem}
\label{theorem:general_obj}
    Let $\betap> 0,\betar> 0$ and $\betap\betar=\beta$. The generalized alignment objective is defined as
    \begin{align}
    \Jlhf^\betar(\pi_\theta^\betap) &= \mathbb{E}_{\x\sim \Datapref}\Big( \mathbb{E}_{\pi_\theta^\betap(\y|\x)}[r_\phi(\x,\y)] \nonumber\\
    &\quad \quad - \betar \KLDiv[\pi_\theta^\betap(\y|\x)\|\pisft(\y|\x)]\Big),
    \label{equ:general_lhf}
\end{align}
    where $\pi_{\theta}^\betap(\y|\x)$ 
    satisfies
    \begin{equation}
    \label{equ:poe}
    \pi_{\theta}^\betap(\y|\x)\propto \pi_{\theta}(\y|\x)^{\betap} \pisft(\y|\x)^{1-\betap}.
\end{equation}
Given unlimited model capacity, the optimal $\pi_{\theta^*}$ that maximizes $\Jlhf^\betar(\pi_\theta^\betap)$ satisfies $\pi_{\theta^*}=\pi_\beta^*$.
\end{restatable}

Intuitively, $\Jlhf^\betar(\pi_\theta^\betap)$ calculates the expectation of the reward regularized with the intensity $\betar$ with respect to the policy regularized with the intensity $\betap$. As the total regularization intensity $\beta=\betar\betap$ is fixed, $\Jlhf^\betar(\pi_\theta^\betap)$ maintains the same analytic solution $\pi_\beta^*$ as $\Jlhf^\beta(\pi_\theta)$. Furthermore, 
it interpolates the policy regularization setting ($\betar=1,\betap=\beta$) in the DPO objective\footnote{$\beta\log\frac{\pi_\theta(\y|\x)}{\pisft(\y|\x)}=\log \frac{\pi_\theta^{\beta}(\y|\x)}{\pisft(\y|\x)} +C(\x)$ where $C(\x)$ depends only on $\x$, which does not effect the DPO objective in Eq. \eqref{equ:dpo}.} and the reward regularization setting ($\betar=\beta,\betap=1$) in the PPO objective when continuously tunning $\betar$ and $\betap$ while keeping their product fixed. We also 
empirically show the effect of $\betar$ and $\betap$ beyond the impact on their product $\beta=\betar\betap$ in Appendix \ref{appendix:ablation_product}.

Next, we derive an equivalent form of the generalized alignment objective by rearranging the elements in Eq. \eqref{equ:general_lhf}, which offers a new insight of the alignment problem from the probability matching perspective. 
The detailed derivation can be found in Appendix \ref{appendix:equivalence}. 
\begin{align}
    \label{equ:lhf_probability_matching}
    \Jlhf^\betar(\pi_{\theta}^\betap)=-\betar &\mathbb{E}_{\x\sim \Datapref}\big[
    \KLDiv(
    \pi_{\theta}^\betap(\y|\x)
    \| \pi^*_\betar(\y|\x)
    )\big] \nonumber\\
    & + \betar \mathbb{E}_{\x\sim \Datapref}\big[\log Z_{\betar}(\x) \big],
\end{align}
As the second term is a constant with respect to $\pi_\theta$, 
Eq. \eqref{equ:lhf_probability_matching} reveals that maximizing the generalized alignment objective $\Jlhf^\betar(\pi_{\theta}^\betap)$ with respect to $\pi_\theta$ is equivalent to minimizing the expected reverse KL divergence $\KLDiv(\pi_{\theta}^\betap\| \pi^*_\betar)$, where $\pi_{\theta}^\betap$ is a function of $\pi_\theta$ as defined in Eq. \eqref{equ:poe}. 

This equivalence implies the possibility of converting the reward maximization problem into a supervised divergence minimization problem, which is able to circumvent the poor stability and low sample efficiency issue caused by high variance in RL solutions~\citep{papini2018stochastic,anschel2017averaged}. 
In the following, we introduce our approach towards exact optimization of this generalized alignment objective by practically realizing the probability matching objective. Without loss of generality, our results remain valid for the original alignment objective in Eq. \eqref{equ:lhf_objective}.


\subsection{Efficient Exact Optimization of the Generalized Alignment Objective}
\label{sec:exo}

We now formally derive EXO
which optimizes the generalized alignment objective $\Jlhf^\betar(\pi_{\theta}^\betap)$ 
by realizing the reverse KL divergence 
$\KLDiv(\pi_{\theta}^\betap\| \pi^*_\betar)$. 
We start with the general setting of language model alignment which preassumes the existence of a 
reward model $r_\phi$; then we consider the case of learning directly from the preference data.

To facilitate policy optimization with straightforward gradient back propagation, we rewrite $\KLDiv(\pi_\theta^\betap \| \pi_{\betar}^*)$ under the expectation of the proposal policy $\pisft$:
\begin{equation}
    \KLDiv(\pi_\theta^\betap \| \pi_{\betar}^*) = \mathbb{E}_{\pisft(\y|\x)}\Bigg[
        \frac{\pi_\theta^\betap(\y|\x)}{\pisft(\y|\x)}
        \log \frac{
            \pi_\theta^\betap(\y|\x)
        }{
            \pi_{\betar}^*(\y|\x)
        }
    \Bigg].\nonumber
\end{equation}
The above expression can be further simplified by defining the log ratio as $f_\theta(\x, \y)=\log\pi_\theta^\betap(\y|\x) - \log{\pisft(\y|\x)}$: 
\begin{equation}
    \KLDiv(\pi_\theta^\betap \| \pi_{\betar}^*) = 
    \mathbb{E}_{\pisft(\y|\x)}\Bigg[
    e^{f_\theta(\x, \y)}
    \log
    \frac{
        e^{f_\theta(\x, \y)}
    }{
        \frac{1}{Z_\betar(\x)}e^{\frac{r_\phi(\x, \y)}{\betar}}
    }
    \Bigg].\nonumber
\end{equation}
As the intractable nature of $Z_\betar(\x)$ hinders direct estimation via Monte Carlo simulation, we propose a practical way to estimate this term by first drawing multiple samples from $\pisft$, and then calculating the reverse KL between the probability distributions defined by ${f_\theta(\x, \y)}$ and ${\frac{1}{\beta_r}r_\phi(\x, \y)}$ over these samples via self-normalization respectively.


\noindent{$\bullet$ \bf Learning from a reward model.} Formally, given $K>1$ i.i.d. completions $\ysamp{1:K}=\{\ysamp{1},\cdots, \ysamp{K}\}$ drawn from $\pisft(\y|\x)$, 
we define an empirical distribution $p_{f_\theta}$ by normalizing the exponential reward $e^{f_\theta(\x, \y)}$ over the $K$ samples:
\begin{equation}
    \label{equ:prob_est_reward}
    p_{f_{\theta}}({i}|\ysamp{1:K}, \x)=\frac{
        e^{f_\theta(\x, \ysamp{i})}
    }{
        \sum_{j=1}^K
        e^{f_\theta(\x, \ysamp{j})}
    }.
\end{equation}
Recall that $f_\theta(\y|\x) = \log\pi_\theta^\betap(\y|\x) - \log{\pisft(\y|\x)}$ and $\pi_{\theta}^\betap(\y|\x)\propto \pi_{\theta}(\y|\x)^{\betap} \pisft(\y|\x)^{1-\betap}$, Eq. \eqref{equ:prob_est_reward} can be rewritten into a form that explicitly depends on $\pi_\theta$:
\begin{equation}
\label{equ:prob_est_policy}
    p_{f_{\theta}}({i}|\ysamp{1:K}, \x)=\frac{
        e^{
            \beta_\pi \log \mfrac{\pi_\theta(\ysamp{i}|\x)}{\pisft(\ysamp{i}|\x)}
        }
    }{
        \sum_{j=1}^K
        e^{
            \beta_\pi \log \mfrac{\pi_\theta(\ysamp{j}|\x)}{\pisft(\ysamp{j}|\x)}
        }
    },
\end{equation}
where $\betap$ can be regarded as the inverse temperature that modulates the empirical distribution defined by the log ratio between $\pi_\theta$ and $\pisft$. Similarly, we define a distribution $p_r$ over the $K$ samples modeled by the reward model $r_\phi$:
\begin{equation}
\label{equ:prob_tgt_policy}
    p_{r_\phi}({i}|\ysamp{1:K},\x) = \frac{
        e^{\frac{1}{\betar}r_\phi(\x, \ysamp{i})}
    }
    {
        \sum_{j=1}^K
        e^{\frac{1}{\betar}r_\phi(\x, \ysamp{j})}
    }.
\end{equation}
Finally, we translate the original objective of reward maximization $\Jlhf^\betar(\pi_{\theta}^\betap)$ into the reverse KL between $p_{f_\theta}$ and $p_{r_\phi}$ over $\ysamp{1:K}$ sampled from $\pisft$:
\begin{align}
\label{equ:exo-objective}
    &\Lexact(\pi_{\theta})=
    \mathbb{E}_{\x\sim \Datapref} \mathbb{E}_{\pisft(\ysamp{1:K}|\x)}
    \Big[\nonumber\\
    &\quad \quad  \KLDiv\big(
    p_{f_\theta}(\cdot|\ysamp{1:K}, \x)\|
    p_{r_\phi}(\cdot|\ysamp{1:K},\x)\big
    )
    \Big].
\end{align}
The complete form of $\Lexact$ is presented in Eq. \eqref{equ:exo_full_form} in Appendix \ref{appendix:proof_exact}. 
Besides its practical simplicity for implementation, we also elucidate its theoretical attributes by characterizing its connection with the generalized alignment objective in Theorem \ref{theorem:rel_lhf}.

\begin{restatable}
{theorem}{ExactFinalTheorem}
\label{theorem:rel_lhf} Following $\pi_\betar^*$, $\pi_\theta^\betap$ and $\Jlhf^\betar(\pi_\theta^\betap)$ defined in Eq. \eqref{equ:optimal}, \eqref{equ:poe}, and \eqref{equ:general_lhf}, for $K\rightarrow\infty$ and arbitary $\theta$, the gradient of $\Lexact(\pi_\theta)$ satisfies
\begin{align}
    \nabla_\theta \Lexact(\pi_\theta) 
    & \!\! = \!\! \nabla_\theta \mathbb{E}_{\x\sim\Datapref}\! \big[
    \KLDiv \!( \pi_\theta^\betap(\y|\x) \| \pi^*_{\betar}(\y|\x)\!)\!\big] \!
    \label{equ:equivalence_rkl} \\
    & \!\! = \! -\frac{1}{\betar} \nabla_\theta \Jlhf^{\betar}(\pi_{\theta}^\betap).
    \label{equ:equivalence_gradient}
\end{align}
\end{restatable}
Theorem \ref{theorem:rel_lhf} suggests the optimization direction for $\pi_\theta$ during the intermediate optimization steps for minimizing $\Lexact(\pi_{\theta})$ aligns with 
the direction required to maximize the generalized alignment objective $\Jlhf^{\betar}(\pi_{\theta}^\betap)$ asymptotically, when sufficient sample population is provided. 
In \S{\ref{sec:synthetic}}, we show that EXO achieves encouraging convergence in practical scenarios with only a finite $K$. Again, as a special case, $\nabla_\theta\Lexact(\pi_\theta)$ aligns with $\nabla_\theta \Jlhf^\beta(\pi_\theta)$ when $\betap=1, \betar=\beta$, which effectively connects with Eq. \eqref{equ:lhf_objective}. The result in Theorem \ref{theorem:rel_lhf} is crucial, as it sheds light on exact optimization of the generalized alignment objective via a simple density matching approach with strong theoretical guarantees. In Appendix \ref{appendix:grad_exo}, we provide a mechanistic understanding of the gradient $\nabla_\theta\Lexact(\pi_\theta)$ which is a weighted sum of the gradients $\nabla_\theta \log\pi_\theta(\y_{k}|\x)$. The weight is proportional to $\KLDiv(p_{f_\theta}\|p_{r_\phi})$ which characterizes the \textit{distributional} gap biased by a \textit{point-wise} correction on the sample $\y_k$.


\noindent{$\bullet$ \bf Learning from human preference data.} In situations where only preference data is accessible, we devise an empirical formulation of $\Lexact$. 
Given a preference dataset $\Datapref$ where each prompt $\x$ is paired with $\y_w$ and $\y_l$ denoting the chosen and rejected completions. This binary supervision can be effectively transformed into an empirical distribution of $p_{\rhuman}$ defined by the underlying reward $\rhuman$ of human preference. 
To avoid infinity when calculating KL divergence, we smooth the one-hot distribution into a soft distribution, i.e., $p_{\rhuman}(w|\y_w, \y_l, \x)= 1-\varepsilon$ and $p_{\rhuman}(l|\y_w, \y_l, \x)=\varepsilon$, where $\varepsilon>0$ is a hyperparameter. $p_{f_\theta}$ can still be computed according to Eq. \eqref{equ:prob_est_policy} over $\y_w$ and $\y_l$. As a result, we present the EXO objective 
on the prefence data by setting $K=2$ and substituting $r_\phi$ with $r_h$ in Eq. \eqref{equ:exo-objective}:
\begin{align}
    \label{equ:exact_pref}
    &\Lexactpref(\pi_{\theta})=
    \mathbb{E}_{(\x,\y_w,\y_l)\sim \Datapref}
    \Big[\nonumber\\
    &\quad \quad \KLDiv\big(
    p_{f_\theta}(\cdot|\y_w, \y_l, \x)\|
    p_{\rhuman}(\cdot|\y_w, \y_l,\x)\big
    )
    \Big].
\end{align}

In practice, $\pisft$ is fine-tuned on either $\Datasft$ which is collected from the same domain as $\Datapref$, or the chosen completions in $\Datapref$ when $\Datasft$ is not available. This closes the distributional gap between $\pisft$ and the unknown distribution that generates the preference data $\Datapref$.

\subsection{Comparing with DPO under the Perspective of Probability Matching}
\label{sec:compare-objs}

Under the perspective of probability matching, we formally demonstrate that the DPO objective corresponds to the \textit{forward} KL which is essentially different from the reverse KL required by the alignment objective $\Jlhf^\beta(\pi_\theta)$ in Eq. \eqref{equ:lhf_objective}. We then analyze their differences under realistic constraints of model capacities.


We first consider the general form of the DPO objective. 
Given $K$ completions $\ysamp{1:K}=\{\ysamp{1},\cdots, \ysamp{K}\}$ drawn from $\pisft(\y|\x)$ and a reward model $r_\phi$ that captures human prefence, we generalize $\Ldpo$ by substituting the sigmoid function with softmax over $K$ responses and replacing the one-hot label 
with a soft distribution defined by $r_\phi$: 
\begin{align}
\label{equ:dpo_rw}
&\Ldpogeneral(\pi_\theta)
=\mathbb{E}_{\x\sim\Datapref}
\mathbb{E}_{
\pisft(\ysamp{1:K}|\x)}\Bigg[ - \sum_{i=1}^K \quad \nonumber \\
    &  \frac{
        e^{\frac{1}{\betar}r_\phi(\x, \ysamp{i})}
    }{  
        \sum_{j=1}^K e^{\frac{1}{\betar}r_\phi(\x, \ysamp{j})}
    }
    \log 
    \frac{
        e^{
            \beta_\pi \log \frac{\pi_\theta(\ysamp{i}|\x)}{\pisft(\ysamp{i}|\x)}
        }
    }{
        \sum_{j=1}^K
        e^{
            \beta_\pi \log \frac{\pi_\theta(\ysamp{j}|\x)}{\pisft(\ysamp{j}|\x)}
        }
    }
    \Bigg].
\end{align}
Upon substituting $r_\phi$ with $\rhuman$ and setting $K=2$, $\Ldpogeneral$ simplifies to $\Ldpo$. 
In the following, we build connection of $\Ldpogeneral(\pi_\theta)$ to the forward KL divegence $\KLDiv(\pi_\betar^* \| \pi_\theta^\betap)$.
\begin{restatable}{theorem}{DPOEquivalenceTheorem}
\label{theorem:dpo_fwkl}
With $\pi_\betar^*$, $\pi_\theta^\betap$ and $p_{r_\phi}$ defined in Eq. \eqref{equ:optimal}, \eqref{equ:poe} and \eqref{equ:prob_tgt_policy} respectively, for $K\rightarrow\infty$ and arbitary $\theta$, the gradient of $\Ldpogeneral(\pi_\theta)$ satisfies
\begin{align}
    \nabla_\theta\Ldpogeneral(\pi_\theta) & =  \nabla_\theta \mathbb{E}_{\x\sim\Datapref}
     \big[ \nonumber \\
     &\KLDiv  (\pi^*_{\betar}(\y|\x) \|
    \pi_\theta^\betap(\y|\x) ) \big].
\end{align}
\end{restatable}
In the following, we consider $\betap=1,\betar=\beta$ to simplify the analysis, while the results still hold for general settings. Putting the results of Theorem \ref{theorem:rel_lhf} and Theorem \ref{theorem:dpo_fwkl} together, we readily connect $\Lexact$ and $\Ldpogeneral$ with two divergences, i.e., the reverse KL, $\KLDiv(\pi_\theta \| \pi^*_\beta)$, and the forward KL, $\KLDiv(\pi^*_\beta \| \pi_\theta)$, respectively. 
To provide a clear image, we illustrate the interrelationship among the objectives of EXO, DPO, and the objective of alignment in Figure \ref{fig:illustrion_equivalency}, as presented in Appendix \ref{appendix:illustration}. Although minimizing these two divergences leads to the same analytic solution $\pi_\beta^*$, they converge to two distinct solutions when considering the expressivity bottleneck of the practical model parametrization.

Specifically, the LM policy $\pi_\theta$ is commonly parametrized as an auto-regressive (AR) model, which enables efficient sampling due to the employment of local normalization. However, the optimal policy $\pi_\beta^*(\y|\x)\propto \pisft(\y|\x)\exp(\frac{1}{\beta}r_\phi(\x,\y))$ defined as an EBM trade-offs sampling efficiency with modeling capacity. Notably, \citet{Lin2021LimitationsOA} rigorously proved that AR models cannot perfectly capture all possible distributions defined by EBMs in terms of supports or rankings based on the computational complexity theories. From an empirical view, this result is also intuitive because the reward model as a 
discriminator is more flexible in distribution modeling than the auto-regressive generator.

Under the practical constraint that $\pi_\theta$ cannot perfectly represent $\pi_\beta^*$, minimizing the forward and reverse KL results in two policies $\pifkl$ and $\pirkl$ with different properties. One well-known fact is that $\pifkl$ is \textit{mean-seeking} while $\pirkl$ is \textit{mode-seeking}~\citep{bishop2006pattern}. 
In Figure \ref{fig:fwkl_revkl}, we consider an illustrative picture of fitting a unimodal $\pi_\theta$ to a multi-modal target $\pi_\beta^*$.
To maintain a minimal forward KL divergence, $\pifkl$ must encompass all the modes of $\pi_\beta^*$ (regions where $\pi_\beta^*$ exhibits significant values). However due to the representational constraints, $\pifkl$ tends to overpresents the mean of different modes of $\pi_\beta^*$, potentially extending into the long tail region of $\pi_\beta^*$~\citep{chan2022greedification,tailr}. On the other hand, $\pirkl$ can select one mode of $\pi_\beta^*$ without causing the reverse KL to explode, meaning that $\pirkl$ will effectively capture the major mode of $\pi_\beta^*$ under realistic model capacity. 
In \S{\ref{sec:synthetic}}, we empirically demonstrate the results of optimizing these two divergences in practice.

Within the context of language model alignment, reverse KL is preferred for generating samples according to the \textit{evaluation} of the optimal policy. Conversely, forward KL tradeoffs preference evaluation with sample diversity, which is rational only if the samples are valid under the evaluation.
To some extent, the reverse KL can also effectively capture this rational diversity, as it maximizes the policy's entropy to prevent distributional collapse to a single point.


\section{Experiments}

We verify the effectiveness of EXO via extensive experimentations. In \S{\ref{sec:synthetic}}, we systematically study the frontier of reward maximization and KL minimization achieved by different alignment methods in a controlled text generation task following previous works~\citep{ziegler_rlhf,rafailov2023direct}. We investigate two different settings, including learning directly from preference data governed by a predefined oracle reward model, and 2) learning from a reward model estimated from the preference data. In \S{\ref{src:real}}, we compare EXO against different approaches on realistic alignment problems including generating human-preferred summaries, helpful dialogue responses, and answers that follow human instructions. Under both settings of learning directly from the preference data and from a reward model, EXO outperforms existing alignment approaches, 
e.g., DPO and PPO. Next, we briefly describe the experiment settings and leave additional details in Appendix \ref{appendix:imp_detail}. 

\vspace{-5pt}
\paragraph{Experiment Setting.} Our experiments consider two practical settings of LM alignment: (i) Directly train on a preference dataset $\Datapref = \{\x^{(j)}, \y_w^{(j)}, \y_l^{(j)}\}_{j=1}^{N}$ where $\y_w$ and $\y_l$ are the chosen and rejected responses judged by an oracle reward model or human labelers. (ii) Train on a reward dataset $\Datarw = \{\x^{(j)}, (\ysamp{1}^{(j)},\rsamp{1}^{(j)}),\cdots,(\ysamp{K}^{(j)},\rsamp{K}^{(j)})\}_{j=1}^N$
where $\ysamp{k}$ is generated by the SFT policy and $r_k$ is a scalar reward provided by a reward model estimated on the given preference dataset. 
In the \textbf{controlled text generation} task, the policy is optimized to generate a completion $\y$ with positive sentiment given a prefix $\x$ of a movie review from the IMDB dataset\footnote{\url{https://huggingface.co/datasets/imdb}.}~\citep{imdb}. To systematically evaluate the alignment performance, we train a binary sentiment classifier on the IMDB dataset and define the oracle reward as its log odds following \citet{ziegler_rlhf}. Both the policy and the reward models are initialized from the GPT-2 large model~\citep{gpt-2}. 
In the \textbf{summarization} task, the policy is required to generate a summary $\y$ of the post $\x$ from the Reddit forum that is preferred by human annotators. Following \citet{summarize_rlhf}, we use the same filtered version\footnote{\url{https://huggingface.co/datasets/UCL-DARK/openai-tldr-filtered}.} of the Reddit TL;DR summarization dataset~\citep{tldr} to train the SFT policy and use their preference dataset\footnote{\url{https://huggingface.co/datasets/openai/summarize_from_feedback}.} for the alignment problem. 
In the \textbf{dialogue generation} task, the policy is learned to generate a helpful response $\y$ given multi-turn dialogue history between the user and the assistant denoted as $\x$. We use the helpfulness subset of the Anthropic Helpful and Harmless dialogue dataset\footnote{\url{https://huggingface.co/datasets/Anthropic/hh-rlhf}.}~\citep{hh} as the preference dataset and train the SFT policy using the chosen responses. 
For summarization and dialogue generation tasks, we initialize both the policy and the reward model from the Pythia-2.8B~\citep{biderman2023pythia} following \citet{rafailov2023direct}. To ensure sample quality, we use a temperature of $\tau=0.8$ to divide the logits of the language model in all experiments. Lastly, for the \textbf{instruction following} task, we create a dataset based on instructions with high demand and representativeness from 
the real-world application scenarios, 
featuring 83K pairs of preferences annotated by human labelers and 49K prompts for policy training. 
The average lengths of the instructions and the answers are 47 and 230 respectively. We curate a diverse set of high-quality test instructions to assess a range of capabilities of the learned LM policy, encompassing multilingual ability, 
creative writing, open-ended question answering, and role playing. Each category takes the same proportion in the test set. Both the policy and the reward models are initialized from ChatGLM2-6B~\citep{glm}. 

\paragraph{Evaluation.} In the controlled text generation task, we evaluate the frontier of the oracle reward and the KL divergence achieved by different approaches. This enables us to systematically compare the effectiveness of different methods in maxmizing the oracle reward under the same distributional shift constrained by the reverse KL. 
For experiments on the public preference datasets of summarization and dialogue generation, we use the reward model trained on the preference dataset as an in-domain proxy of the unknown ground-truth reward and also query GPT-4 for zero-shot pair-wise evaluation, which is shown to be consistent with human judgments~\citep{rafailov2023direct}. The prompts for GPT-4 evaluation are slightly modified based on those used in \citet{rafailov2023direct}, as detailed in Appendix \ref{appendix:imp_detail}. We compare the generated outputs against those generated by the SFT policy and the preferred choice in the preference dataset. For the instruction-following task, we report the win rate of directly comparing our method against various baselines as judged by GPT-4. Additionally, we employ human assessment to evaluate criteria including adherence to instruction, correctness, fluency, safety and helpfulness.

\paragraph{Methods for Comparison.} 
We consider the following methods for aligning language models with human preferences under various settings. 
Under the setting of learning directly from preferences, we consider the special case of {EXO} for preference learning $\Lexactpref$ (Eq. \eqref{equ:exact_pref}) denoted as \textbf{\EXOpref}, and the standard DPO~\citep{rafailov2023direct} that minimizes the pair-wise loss $\Ldpo$ (Eq. \eqref{equ:dpo}) on the preference data, which we denoted as \textbf{\DPOpref}. 
Then we consider the setup of alignment with a reward model estimated from the preference dataset, which includes the RL algorithm \textbf{PPO}~\citep{ziegler_rlhf} that optimizes the expected reward with a KL penalty (Eq. \eqref{equ:lhf_objective}), the general EXO objective $\Lexact$ (Eq. \eqref{equ:exo-objective}) that performs probability matching by minimizing reverse KL, which is denoted as \textbf{\EXOrw}, the general DPO objective $\Ldpogeneral$ (Eq. \eqref{equ:dpo_rw}) that minimizes the forward KL, which is denoted as \textbf{\DPOrw}, and the \textbf{Best-of-$N$} method which first samples $N=128$ outputs from the SFT policy and then returns the response with the highest score according to the reward model. Note that the Best-of-$N$ baseline is practically inefficient and can be regarded as an upperbound of exploiting the SFT policy according to the reward model in Eq. \eqref{equ:optimal} by trading off the computation. 

\subsection{Alignment with the Oracle Reward}
\label{sec:synthetic}

\begin{figure}[t!]
    \centering
    \includegraphics[width=0.85\columnwidth]{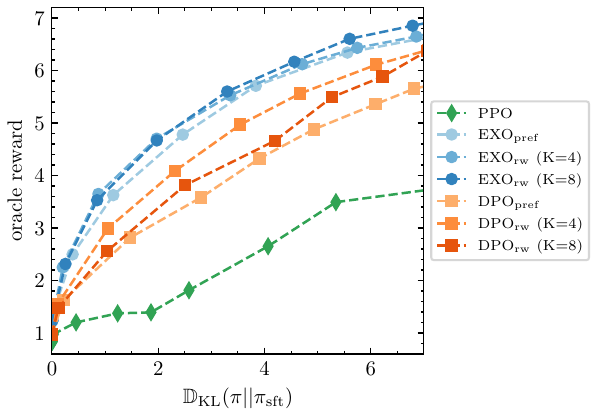}
    \vspace{-7pt}
    \caption{The frontier of oracle reward vs reverse KL to the SFT policy of different methods in the controlled experiment.
    \vspace{-15pt}
    }
    \label{fig:imdb_rw_kl}
\end{figure}

To avoid undesirable reward overoptimizaion due to distributional shift, a preferred alignment solution should return a policy that obtains high oracle reward while incuring minimum deviation from $\pisft$. 
Thereby, we plot the frontier of the oracle reward against KL divergence in Figure \ref{fig:imdb_rw_kl}. We additionally present the accuracy-KL frontier in Figure \ref{fig:imdb_acc_kl} in Appendix \ref{appendix:imdb_acc_kl}. Each point represents a checkpoint of the learned policy which is evaluated on 512 prefixes from the test set to complete the response with maximumly 512 tokens. We sample $M=4$ completions $\{\y_i\}_{i=1}^M$ for each given prompt $\x$ to calculate the average oracle reward as well as to reduce the variance of approximating the sequence-level KL divergence $\KLDiv(\pi_\theta\|\pisft) \approx \frac{1}{M}\sum_{i=1}^M \log \pi_\theta(\y_i|\x) - \log \pisft(\y_i|\x)$. 

Despite aiming to optimize the same alignment objective, the EXO approaches ({\EXOpref} and {\EXOrw}) yield the most efficient frontiers in their respective settings
, evidenced by consistently achieving higher oracle rewards than baselines under the same KL divergence. Specifically, in the setting of directly learning from preference pairs, {\EXOpref} outperforms {\DPOpref} by a large margin, which underscores EXO's better sample efficiency when learning from a constrained source of preference indicators. As $K$ increases, the frontier of {\EXOrw} begins to exhibit convergence, indicating an effective progression towards the intended solution within a finite $K$. 
Although {\DPOrw} also improves over {\DPOpref} when $K=4$, the frontier becomes worse when $K$ is further increased to 8. This result substantiates our analysis about the mean-seeking behavior of forward KL, which leads to inaccuracy in capturing the modes of the complex target distribution. 
Finally, we illustrate the strong optimization efficiency of EXO in Figure \ref{fig:orc_vs_steps}, evidenced by consistently achieving high and stable oracle rewards within fewer number of training steps compared with PPO and DPO in Appendix \ref{appendix:efficiency}. 

\begin{figure}
    \centering
    \includegraphics[width=0.95\columnwidth]{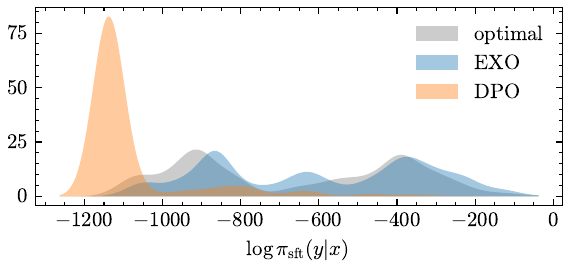}
    \vspace{-10pt}
    \caption{Visualization of the estimated density ratio between the optimal and learned policy by EXO and DPO and the SFT policy on samples from the SFT policy sorted by their log probabilities.}
    \label{fig:imdb_vis}
    \vspace{-5pt}
\end{figure}

Next, we compare DPO and EXO from the probability matching perspective by visualizing the probability density of the policies obtained by these two approaches\footnote{We consider the setting of learning directly from preferences.}. 
In Figure \ref{fig:imdb_vis}, we plot the estimated density ratio of the optimal and learned policies by EXO and DPO against $\pisft$ given a randomly chosen test prompt ``\textit{This Fox spectacle was a big hit when released in}''. Since the probability density of an LM policy is defined over a high dimensional space of $\Y=\V^n$, it is intractable to evaluate every point in this space exhaustively. Thus, we consider the representative data points that are sampled from $\pisft$, and sort them in the ascending order of their log probabilities. 
Then we compute the empirical distribution under the learned policies over these samples. Formally, given $M=256$ samples $\{\y_i\}_{i=1}^M$ drawn from $\pisft$ conditioned on the prompt $\x$, the empirical distribution $\hat{\pi}_\theta$ is calculated via self-normalized importance sampling over the learned policy $\pi_\theta$:
\begin{equation*}
    \hat{\pi}_\theta(\y_i|\x)= \frac{M\pi_\theta(\y_i|\x)}{\sum_{j=1}^M \pi_\theta(\y_j|\x)/\pisft(\y_j|\x)}.
\end{equation*}
For the optimal policy, the empirical distribution reduces to: 
\begin{equation*}
    \hat{\pi}^*_\beta(\y_i|\x) = \frac{M\pisft(\y_i|\x)\exp({r(\x, \y_i)/\beta})}{\sum_{j=1}^M \exp({r(\x, \y_j)/\beta})}.
\end{equation*}
Finally, we use kernel density estimation to estimate the probability density $\hat{\pi}(\y|\x)$ of the empirical distribution and plot the density ratio $\rho_{\hat{\pi}}(\y|\x) = \frac{\hat{\pi}(\y|\x)}{\pisft(\y|\x)}$ against the log probability of the data points under $\pisft$. From the result, the density ratio of the EXO policy closely matches the optimal policy at the high probability region against $\pisft$, which reflects its ability in capturing the major modes of the optimal policy. However, the DPO policy overestimates the long tail in $\pisft$ due to the mean-seeking tendency of forward KL, resulting in a reduced accuracy 
in fitting the modes of the optimal policy. We present more visualization results in Appendix \ref{appendix:vis}.


\subsection{Alignment with Human Preferences}
\label{src:real}

\begin{table}[t!]
    \centering
    \small
    \begin{tabular}{lcccc}
    \toprule
    \multirow{2}{*}{Method} & \multicolumn{2}{c}{Reward Model (\%)} & \multicolumn{2}{c}{GPT-4 (\%)} \\
    & vs SFT & vs Chosen & vs SFT & vs Chosen \\
    \midrule
    \multicolumn{5}{c}{\rowfont{\footnotesize}\texttt{w/ Preferences}} \\
    \midrule
    \DPOpref & 68.3 & 23.7 & 57.0 & 30.5 \\
    \EXOpref & \textbf{92.5} & \textbf{60.1} & \textbf{83.0} & \textbf{55.0} \\
    \midrule
    \multicolumn{5}{c}{\rowfont{\footnotesize}\texttt{w/ Reward Model}} \\
    \midrule
    Best-of-$N$ & {99.3} & 75.8 & 83.5 & 60.0 \\
    PPO & 93.2 & 58.3 & 77.0 & 52.0 \\
    \DPOrw & 82.7 & 39.8 & 70.0 & 41.0 \\
    \EXOrw & \textbf{97.3} & \textbf{76.4} & \textbf{88.5} & \textbf{64.0} \\
    \bottomrule
    \end{tabular}
    \caption{Win rates against the SFT generated texts and the chosen texts on the TL;DR summarization dataset. Best results from the computationally efficient methods are highlighted in \textbf{boldface}.
    \vspace{-10pt}}
    \label{tab:tldr}
\end{table}

Next, we undertake a series of experiments focused on 
learning from real human preferences.
In the tasks of summarization and dialogue generation, we consider the two settings of learning directly from preference data and from a reward model respectively. We set the same hyperparameters (e.g., $\betap$, $\betar$) for EXO and DPO across different settings and datasets, and provide the results of tuning these hyperparameters in Appendix \ref{appendix:ablation} and \ref{appendix:ablation_product}. 
Evaluation results on text summaization and dialogue generation are shown in Table \ref{tab:tldr} and Table \ref{tab:hh} respectively. Upon comparison with both the SFT and chosen responses, EXO demonstrates notable improvement over DPO and PPO. This advantage is evident in evaluations using both the in-domain reward model and zero-shot assessment by GPT-4, across both settings of learning from preferences and from a reward model. 
Notably, EXO is the only practically efficient method to attain a GPT-4 win rate exceeding 60\% when compared to the chosen responses that may have been produced by a more advanced language model. Although the Best-of-$N$ baseline achieves comparable or higher reward model win rate by maximizing out the reward model, it suffers from the most significant decline of win rate when assessed by GPT-4. This drop of performance is attributed to its excessive exploitation of the imperfect reward model while neglecting the deviation from the initial SFT policy. We provide examples generated by DPO and EXO on both tasks in Appendix \ref{appendix:sample_outputs}.

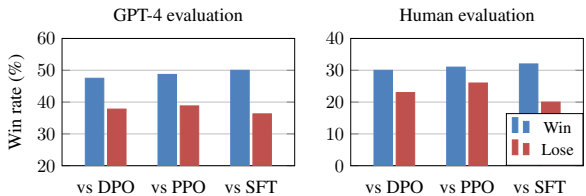
\begin{figure}[h!]
\vspace{-5pt}
    \centering
    \subcaptionbox{}{
        \scalebox{0.7}{\begin{tikzpicture}

\begin{axis} 
[ybar,
height=4cm,
width=6cm,
major x tick style = transparent,
bar width=10pt,
ymajorgrids = true,
ylabel = {Win rate (\%)},
symbolic x coords={{vs DPO}, 
{vs PPO}, 
{vs SFT}
},
xtick = data,
scaled y ticks = false,
enlarge x limits=0.3,
ymin=20,
ymax=60,
title=GPT-4 evaluation
]

\addplot[style={bblue,fill=bblue,mark=none}]
            coordinates {
            ({vs DPO}, 47.5) 
            ({vs PPO}, 48.7)
            ({vs SFT},50.0)
            };

\addplot[style={rred,fill=rred,mark=none}]
             coordinates {
             ({vs DPO}, 37.8) 
             ({vs PPO}, 38.8)
             ({vs SFT},36.3)
             };

\end{axis}
 
\end{tikzpicture}}
        \phantomsubcaption
    }
    \subcaptionbox{}{
        \scalebox{0.7}{\begin{tikzpicture}

\begin{axis} 
[ybar,
height=4cm,
width=6cm,
major x tick style = transparent,
bar width=10pt,
ymajorgrids = true,
symbolic x coords={{vs DPO}, 
{vs PPO}, 
{vs SFT}
},
xtick = data,
scaled y ticks = false,
enlarge x limits=0.3,
ymin=0,
ymax=40,
title=Human evaluation,
legend style={
                at={(1.0,0)},
                anchor=south east,
                column sep=1ex,
                font=\small
        }
]
\addplot[style={bblue,fill=bblue,mark=none}]
            coordinates {
            ({vs DPO}, 30.0) 
            ({vs PPO}, 31.0)
            ({vs SFT},32.0)
            };

\addplot[style={rred,fill=rred,mark=none}]
             coordinates {
             ({vs DPO}, 23.0) 
             ({vs PPO}, 26.0)
             ({vs SFT},20.0)
             };

\legend{Win,Lose}
\end{axis}
 
\end{tikzpicture}}
    }
    \vspace{-15pt}
    \caption{
    Win rates by comparing EXO to various baselines on the instruction-following task judged by GPT-4 and human labelers.
    \vspace{-10pt}}
    \label{fig:instruct_exp}
\end{figure}

For the instruction-following task, we report the win rates of EXO compared to various baselines in Figure \ref{fig:instruct_exp} under the setting of learning from the reward model given its advantageous performance observed so far. From the result, we observe that EXO outperforms all baselines by clear margins, thereby underscoring its scalability in practical applications. Notably, EXO achieves 10\% and 5\% improvement over its closest competitors as judged by GPT-4 and human annotators respectively.

\begin{table}[t!]
    \centering
    \small
    \begin{tabular}{lcccc}
    \toprule
    \multirow{2}{*}{Method} & \multicolumn{2}{c}{Reward Model (\%)} & \multicolumn{2}{c}{GPT-4 (\%)} \\
    & vs SFT & vs Chosen & vs SFT & vs Chosen \\
    \midrule
    \multicolumn{5}{c}{\rowfont{\footnotesize}\texttt{w/ Preferences}} \\
    \midrule
    \DPOpref & 66.3 & 65.1 & 58.0 & 37.0 \\
    \EXOpref & \textbf{76.4} & \textbf{76.7} & \textbf{73.0} & \textbf{51.0}\\
    \midrule
    \multicolumn{5}{c}{\rowfont{\footnotesize}\texttt{w/ Reward Model}} \\
    \midrule
    Best-of-$N$ & 94.6 & 98.2 & 86.0 & 63.0 \\
    PPO & 75.0 & 74.0 & 66.5 & 52.0 \\
    \DPOrw & 79.9 & 81.3 & 75.5 & 49.0\\
    \EXOrw & \textbf{85.6} & \textbf{87.2} & \textbf{83.5} & \textbf{60.0} \\
    \bottomrule
    \end{tabular}
    \caption{Win rates against the SFT generated texts and the chosen texts on the Anthropic-HH dataset. Best results from the computationally efficient methods are highlighted in \textbf{boldface}.
    \vspace{-10pt}}
    \label{tab:hh}
\end{table}

\captionsetup[subfigure]{labelformat=empty}
\section{Related Work}

Large language models~\citep{gopher,chinchila,touvron2023llama} learned to predict the next tokens on large corpora have significantly improved the performance of various natural language processing (NLP) tasks in zero shot~\citep{gpt-2} or few-shot~\citep{gpt-3,palm} settings. To bridge the mismatch between the training objective and users' objective of solving specific NLP tasks, instruction-tuning is proposed, which fine-tunes the language models on human curated instruction-response pairs in a fully supervised manner~\citep{sanh2021T0,chung2022scaling,longpre2023flan,self-instruct}. Despite its performance and generalization to unseen instruction sets~\citep{chung2022scaling}, there have been concerns that the model learned to predict the next token might only capture surficial patterns rather than comprehending the tasks~\citep{kung2023models}. 

To address the aforementioned deficiencies, the framework of reinforcement learning from human feedback (RLHF) is proposed~\citep{ziegler_rlhf}, which relies on only \textit{relative} human preferences on response quality and optimizes the language model by RL algorithms~\citep{REINFORCE}, e.g., PPO~\citep{schulman2017proximal}, under the supervision of a reward model which encapsulates the implicit preference of humans. Despite the success of RLHF in various tasks where alignment is strongly emphasized, e.g., translation~\citep{kreutzer2018reliability}, summarization~\citep{summarize_rlhf}, instruction following~\citep{instruct_gpt}, and etc., RL fine-tuning for large language models still faces serious challenges in stability and scalability~\citep{zheng2023secrets}. 

Thereby, a new trend of recent works proposed to optimize the language model to follow human preferences without applying RL algorithms~\citep{yuan2023rrhf,zhao2023slic,rafailov2023direct}. While most approaches are empirically set to model alignment as a re-ranking problem, DPO proposed a theoretically sound way to realize direct policy optimization from preference data based on the analytic optimal solution of the reward maximization objective in RLHF. Afterwards, there have been several studies that extend DPO in various ways, e.g., altering the preference data by sampling from the optimal policy via rejection sampling~\citep{liu2023statistical}, substituting the point-wise reward with a pair-wise preference function~\citep{ipo}, extending the preference pairs to rankings of preferences of any size~\citep{Song2023PreferenceRO}, and etc. However, these approaches are still based on the formation of DPO. In this work, we propose to learn the policy by exactly optimizing the RLHF objective via probability matching that minimizes the reverse KL. In the literature of maximum-entropy RL~\citep{MaxEntRLQuestion,RLBayesianInference}, this equivalent form of probability matching was discussed only for analysis purposes, while we are the first to derive a practical way to optimize it. We also revisit DPO under this perspective and recognize that it actually corresponds to minimizing the forward KL in its general form. To this sense, DPO shares the same spirit of weighted regression~\citep{reward-wr,WeightedRegression,DBLP:conf/nips/0001NZMSRSSGHF20}, an algorithm that directly utilizes behavioral actions to supervise the policy in offline RL~\citep{DBLP:conf/aaai/PetersMA10,cep}. However, this approach is known to be suboptimal when the policy model is limited in distributional
expressivity~\citep{yue2022boosting,DBLP:conf/iclr/Chen0Y0023}. We analyze the characteristics of the probability density learned by DPO with both theoretical insight and empirical experimentations.

\section{Limitations and Future Work}

The alignment framework proposed by \citet{ziegler_rlhf} relies on the KL regularization to the SFT policy to prevent the optimized policy from greedily maximizing out the reward model which is estimated from the human preference data. Despite the regularization, our experiments still reveal instances of reward over-optimization, a phenomenon possibly due to insufficient focus on the reward model estimation and use. Importantly, ensuring that the reward model accurately reflects the true modes of the oracle distribution of human is more vital for PPO and EXO that optimizes a mode-seeking objective. This opens possible avenues to improve the alignment framework at a broader scope. For instance, rather than relying on a static preference dataset for reward model training, it could be more effective to dynamically improve the reward model with the development of the policy, thereby offering more precise feedback. Upon the current regularization that solely focuses on proximity to the initial policy, one can take into account the uncertainty of the reward model output to avoid over-exploitation of the reward model. It is beneficial to take into account these aspects to develop efficient and effective method towards closer alignment with human preferences. Additionally, while we already evaluated EXO on advanced language models up to 6B on realistic scenarios, scaling EXO to models that are orders of magnitude larger can present profound implications. At the other end of the spectrum, systematically dissecting and comparing PPO, DPO and EXO, e.g., regarding their variance and bias during optimization is essential to broaden our understanding of these methods.

\section{Conclusion}

In this work, we consider the problem of aligning language models with human preferences. Although reinforcement learning (RL) for reward maximization presents the direct and apparent solution, we reframe the problem in a supervised probability matching framework, which underscores the probabilistic interpretation of the alignment procedure. 
This derives our \textit{efficient exact optimization} (EXO) of the KL-regularized reward maximization objective of alignment. Formally, we prove the asymptotic equivalence between the EXO objective and the alignment objective. In practice, EXO enables efficient optimization via probability matching between empirical distributions, 
which avoids the complexities of RL algorithms. 
We further demonstrate that DPO in its general form actually corresponds to minimizing the forward KL against the optimal policy, which is shown to be less effective in capturing the modes of the optimal policy under realistic model parametrization with both theoretical and empirical justifications. Finally, we demonstrate the effectiveness and scalability of EXO on various text generation tasks with real human preferences.

\arxivornot{
    
    \section*{Acknowledgements}

    This work was supported by the NSFC projects (with No. 61936010 and No. 62306160). This work was supported by the National Science Foundation for Distinguished Young Scholars (with No. 62125604). This work was also supported by China National Postdoctoral Program for Innovative Talents (No. BX20230194) and China Postdoctoral Science Foundation (No. 2023M731952). We would also like to thank Zhipu AI for sponsoring the computation resources and annotation cost used in this work.

    \section*{Impact Statement}
    This paper presents a method whose goal is to advance the alignment of language models with human preferences. This endeavor, while technically challenging, carries significant implications for the ethical use and societal impact of artificial intelligence. The goal of alignment aims to mitigate the inherent biases of AI systems and ensure that they reflect diverse human values, goals and intentions that are safe and ethical. It enhances AI's utility in various sectors to make reliable decisions that are in line with organizational goals and ethical standards. 
}
{
    \section*{Impact Statement}

    This paper presents a method whose goal is to advance the alignment of language models with human preferences. This endeavor, while technically challenging, carries significant implications for the ethical use and societal impact of artificial intelligence. The goal of alignment aims to mitigate the inherent biases of AI systems and ensure that they reflect diverse human values, goals and intentions that are safe and ethical. It enhances AI's utility in various sectors to make reliable decisions that are in line with organizational goals and ethical standards. 
}

\bibliography{icml2024}
\bibliographystyle{icml2024}

\newpage
\appendix
\onecolumn

\section{Proofs and Derivations}
\label{appendix:proofs}

\subsection{Proof of Theorem \ref{theorem:general_obj}}
\label{appendix:optimal_altern_lhf}
\begin{proof}
We derive the optimal $\pi_{\theta^*}$ that maximizes the generalized alignment objective $\Jlhf^{\betar}(\pi_\theta^{\betap})$ and show that it equals to the optimal solution $\pi_\beta^*$ of $\Jlhf^\beta(\pi_\theta)$ given unlimited model capacity. First, we restate the formation of $\Jlhf^{\betar}(\pi_\theta^{\betap})$:
\begin{equation}
    \label{equ:alternative_lhf_appendix}
    \Jlhf^\betar(\pi_\theta^\betap) = \mathbb{E}_{\x\sim \Datapref}\Big( \mathbb{E}_{\pi_\theta^\betap(\y|\x)}[r_\phi(\x,\y)] - \betar \KLDiv[\pi_\theta^\betap(\y|\x)\|\pisft(\y|\x)]\Big),
\end{equation}
where $\pi_\theta^\betap$ is defined as:
\begin{equation}
    \label{equ:poe_appendix}
    \pi_\theta^\betap(\y|\x) \propto \pi_\theta(\y|\x)^{\betap} \pisft(\y|\x)^{1-\betap}.
\end{equation}
Then we substitute $\theta$ with the optimal $\theta^*$ in Eq. \eqref{equ:poe_appendix} where $\pi_{\theta^*}$ maximizes $\Jlhf^\betar(\pi_\theta^\betap)$:
\begin{equation}
    \label{equ:poe_optimal_appendix}
    \pi_{\theta^*}^\betap(\y|\x) \propto \pi_{\theta^*}(\y|\x)^{\betap} \pisft(\y|\x)^{1-\betap}.
\end{equation}
Since $\pi_{\theta^*}^\betap$ is also the optimal policy that maximizes $\Jlhf^\betar(\cdot)$, it should satisfy Eq. \eqref{equ:optimal} which gives:
\begin{equation}
    \label{equ:optimal_betar}
    \pi_{\theta^*}^\betap(\y|\x)=\pi_{\betar}^*(\y|\x)\propto \pisft(\y|\x)e^{\frac{1}{\betar}r(\x,\y)}.
\end{equation}

Together with Eq. \eqref{equ:poe_optimal_appendix} and Eq. \eqref{equ:optimal_betar}, we obtain the form of $\pi_{\theta^*}$ via some simple algebra:
\begin{align*}
    \pi_{\theta^*}(\y|\x) & \propto (\pi_\betar^*(\y|\x)\pisft(\y|\x)^{\betap - 1}) ^{\frac{1}{\betap}} \\
    & \propto (\pisft(\y|\x)e^{\frac{1}{\betar}r(\x, \y)})^{\frac{1}{\betap}} \pisft(\y|\x)^{\frac{\betap - 1}{\betap}}\\
    & \propto \pisft(\y|\x)^{\frac{1}{\betap}}e^{\frac{1}{\betar\betap}r(\x, \y)}\pisft(\y|\x)^{\frac{\betap - 1}{\betap}}\\
    & \propto \pisft(\y|\x) e^{\frac{1}{\betar\betap}r(\x, \y)}.
\end{align*}
By definition, $\betar\betap=\beta$, then $\pi_{\theta^*}$ reduces to the same form of the optimal solution of $\Jlhf^\beta(\pi_\theta)$ defined in Eq. \eqref{equ:optimal}:
\begin{equation*}
    \pi_{\theta^*}(\y|\x) = \pi_{\beta}^*(\y|\x) \propto \pisft(\y|\x) e^{\frac{1}{\beta}r(\x, \y)},
\end{equation*}
which completes the proof.
\end{proof}

\subsection{Derivation of Eq. \eqref{equ:lhf_probability_matching}}
\label{appendix:equivalence}

We first start by rearranging $\Jlhf^{\betar}(\pi_\theta^{\betap})$ into the expectation of a log ratio:
\begin{align*}
    \Jlhf^\betar(\pi_\theta^\betap) 
    &= \mathbb{E}_{\x\sim \Datapref}\Big( \mathbb{E}_{\pi_\theta^\betap(\y|\x)}[r_\phi(\x,\y)] - \betar \KLDiv[\pi_\theta^\betap(\y|\x)\|\pisft(\y|\x)]\Big) \\
    &=\mathbb{E}_{\x\sim \Datapref}\Big( \mathbb{E}_{\pi_\theta^\betap(\y|\x)}[r_\phi(\x,\y)]
    -\betar \mathbb{E}_{\pi_\theta^\betap(\y|\x)}
    \Big[
    \log \frac{
        \pi_\theta^\betap(\y|\x)
    }{
        \pisft(\y|\x)
    }
    \Big]
    \Big)\\
    &=\mathbb{E}_{\x \sim \Datapref}
    \Big(
    \betar\mathbb{E}_{\y \sim \pi_\theta^\betap(\y|\x)}
    \big[
    \log e^{\frac{1}{\betar} r_\phi(\x, \y)}
    \big]
    - \betar \mathbb{E}_{\pi_\theta^\betap(\y|\x)}
    \Big[
    \log \frac{
        \pi_\theta^\betap(\y|\x)
    }{
        \pisft(\y|\x)
    }
    \Big]
    \Big)\\
    &=\mathbb{E}_{\x \sim \Datapref}\mathbb{E}_{\y \sim \pi_\theta^\betap(\y|\x)}
    \bigg[
    \betar \log 
    \frac{
        \pisft(\y|\x)e^{\frac{1}{\betar} r_\phi(\x, \y)}
    }{
        \pi_\theta^\betap(\y|\x)
    }
    \bigg]
\end{align*}

Notice the analytical form of $\pi_\betar^*$:
\begin{equation*}
    \pi_\betar^*(\y|\x)=
    \frac{1}{Z_{\betar(\x)}}\pisft(\y|\x)e^{\frac{1}{\betar}r_\phi(\x,\y)}.
\end{equation*}
We substitute $\pisft(\y|\x)e^{\frac{1}{\betar}r_\phi(\x,\y)}$ into the expression of $\Jlhf^{\betar}(\pi_\theta^{\betap})$:
\begin{align*}
    \Jlhf^\betar(\pi_\theta^\betap) 
    &=\mathbb{E}_{\x \sim \Datapref}\mathbb{E}_{\y \sim \pi_\theta^\betap(\y|\x)}
    \bigg[
        \betar \log 
        \frac{
            Z_{\betar(\x)} \pi_\betar^*(\y|\x)
        }{
            \pi_\theta^\betap(\y|\x)
        }
    \bigg]\\
    &=\betar \mathbb{E}_{\x\in\Datapref}\Big[
        -\KLDiv(\pi_\theta^\betap(\cdot|\x) \| \pi_\betar^*(\cdot|\x))
        + \log Z_\betar(\x)
    \Big].
\end{align*}

\subsection{Proof of Theorem \ref{theorem:rel_lhf}}
\label{appendix:proof_exact}

\begin{proof}
We first restate the definition of $\Lexact(\pi_\theta)$ by substituting Eq. \eqref{equ:prob_est_reward}, \eqref{equ:prob_tgt_policy} into \eqref{equ:exo-objective}:
\begin{equation}
    \label{equ:exo_full_form}
    \Lexact(\pi_\theta) = 
    \mathbb{E}_{\x\sim \Datapref}
    \mathbb{E}_{\pisft(\y_{1:K}|\x)}
    \Bigg[
    \sum_{i=1}^K
    \frac{
        e^{
            f_\theta(\x, \y_i)
        }
    }{
        \sum_{j=1}^K
        e^{
             f_\theta(\x, \y_j)
        }
    } 
    \Big(
    \log 
    \frac{
        e^{
             f_\theta(\x, \y_i)
        }
    }{
        \sum_{j=1}^K
        e^{
             f_\theta(\x, \y_j)
        }
    }
    - \log 
    \frac{
        e^{\frac{1}{\betar} r_\phi(\x, \y_i)}
    }{
        \sum_{j=1}^K
        e^{\frac{1}{\betar} r_\phi(\x, \y_j)}
    }
    \Big)
    \Bigg]
\end{equation}
Since $f_\theta(\x, \y)=\log \pi_\theta^\betap(\y|\x) - \log \pisft(\y|\x)$, we have that:
\begin{align}
    \Lexact(\pi_\theta) &= \nonumber
    \mathbb{E}_{\x\sim \Datapref}
    \mathbb{E}_{\pisft(\y_{1:K}|\x)}
    \Bigg[
    \sum_{i=1}^K
    \frac{
        e^{
            \log \frac{
                \pi_\theta^\betap(\y_i|\x)    
            }{
                \pisft(\y_i|\x)    
            }
        }
    }{
        \sum_{j=1}^K
        e^{
             \log \frac{
                \pi_\theta^\betap(\y_j|\x)    
            }{
                \pisft(\y_j|\x)    
            }
        }
    } 
    \Big(
    \log 
    \frac{
        e^{
             \log \frac{
                \pi_\theta^\betap(\y_i|\x)    
            }{
                \pisft(\y_i|\x)    
            }
        }
    }{
        \sum_{j=1}^K
        e^{
             \log \frac{
                \pi_\theta^\betap(\y_j|\x)    
            }{
                \pisft(\y_j|\x)    
            }
        }
    }
    - \log 
    \frac{
        e^{\frac{1}{\betar} r_\phi(\x, \y_i)}
    }{
        \sum_{j=1}^K
        e^{\frac{1}{\betar} r_\phi(\x, \y_j)}
    }
    \Big)
    \Bigg] \\
    &=
    \mathbb{E}_{\x\sim \Datapref}
    \mathbb{E}_{\pisft(\y_{1:K}|\x)}
    \Bigg[
    \sum_{i=1}^K
    \frac{
        \frac{
                \pi_\theta^\betap(\y_i|\x)    
            }{
                \pisft(\y_i|\x)    
            }
    }{
        \sum_{j=1}^K
        \frac{
                \pi_\theta^\betap(\y_j|\x)    
            }{
                \pisft(\y_j|\x)    
            }
    } 
    \Big(
    \log 
    \frac{
        \frac{
                \pi_\theta^\betap(\y_i|\x)    
            }{
                \pisft(\y_i|\x)    
            }
    }{
        \sum_{j=1}^K
        \frac{
                \pi_\theta^\betap(\y_j|\x)    
            }{
                \pisft(\y_j|\x)    
            }
    }
    - \log 
    \frac{
        e^{\frac{1}{\betar} r_\phi(\x, \y_i)}
    }{
        \sum_{j=1}^K
        e^{\frac{1}{\betar} r_\phi(\x, \y_j)}
    }
    \Big)
    \Bigg].
    \label{equ:exo_intermediate}
\end{align}
Since $\{\y_i\}_{i=1}^K$ are sampled from $\pisft(\cdot|\x)$, when $K\rightarrow\infty$, for arbitary function $g: \Y \rightarrow \mathbb{R}$, the estimate $\frac{1}{K}\sum_{i=1}^K g(\y_i)$ is unbiased, i.e., $\lim_{K\rightarrow\infty}\frac{1}{K}\sum_{i=1}^K g(\y_i)=\mathbb{E}_{\pisft(\y|\x)}[g(\y)]$. We consider the following two instantiations of $g(\cdot)$.

For $g(\y) = \frac{
                \pi_\theta^\betap(\y|\x)    
            }{
                \pisft(\y|\x)    
            }$, we have:
\begin{align*}
    \sum_{j=1}^K 
    \frac{
                \pi_\theta^\betap(\y_j|\x)    
            }{
                \pisft(\y_j|\x)    
            }
    &= K \mathbb{E}_{\pisft(\y|\x)}
    \bigg[    
    \frac{\pi_\theta^\betap(\y|\x)}
    {\pisft(\y|\x)}
    \bigg]\\
    &=K\sum_{\y \in \Y} \pisft(\y|\x)
    \frac{\pi_\theta^\betap(\y|\x)}
    {\pisft(\y|\x)}\\
    &=K\sum_{\y \in \Y} \pi_\theta^\betap(\y|\x)\\
    &=K.
\end{align*}

For $g(\y) = e^{\frac{1}{\betar}r_\phi(\x, \y)}$, we have:
\begin{align*}
    \sum_{j=1}^K e^{\frac{1}{\betar}r_\phi(\x, \y_j)} & = K \mathbb{E}_{\pisft(\y|\x)}\Big[
    e^{\frac{1}{\betar}r_\phi(\x, \y)}
    \Big] \\
    & = K \sum_{\y\in\Y}\pisft(\y|\x)
    e^{\frac{1}{\betar}r_\phi(\x, \y)}\\
    &=K Z_{\betar}(\x).
\end{align*}

Then we simplify $\Lexact$ by substituting the expression of $\sum_{j=1}^K 
\frac{
            \pi_\theta^\betap(\y_j|\x)    
        }{
            \pisft(\y_j|\x)    
}$ and $\sum_{j=1}^K e^{\frac{1}{\betar}r_\phi(\x, \y_j)}$ when $K\rightarrow\infty$ into Eq. \eqref{equ:exo_intermediate}.
\begin{align*}
\Lexact(\pi_\theta) &= 
\mathbb{E}_{\x\sim \Datapref}
    \mathbb{E}_{\pisft(\y_{1:K}|\x)}
    \Bigg[
    \sum_{i=1}^K 
    \frac{
            \pi_\theta^\betap(\y_i|\x)    
        }{
            K\pisft(\y_i|\x)    
    }
    \Big(
    \log 
    \frac{
            \pi_\theta^\betap(\y_i|\x)    
        }{
            K\pisft(\y_i|\x)    
    }
    -
    \log \frac{
        e^{\frac{1}{\betar}r_\phi(\x, \y_i)}
    }{KZ_{\betar}(\x)}
    \Big)
    \Bigg] \\
    &= 
    \mathbb{E}_{\x\sim \Datapref}
    \mathbb{E}_{\pisft(\y_{1:K}|\x)}
    \Bigg[
    \frac{1}{K}
    \sum_{i=1}^K
    \frac{\pi_\theta^\betap(\y_i|\x)}{\pisft(\y_i|\x)}
    \log
    \frac{
        \pi_\theta^\betap(\y_i|\x)
    }{
        \pisft(\y_i|\x)
        {
            e^{\frac{1}{\betar} r_\phi(\x, \y_i)}
        }/{
            Z_\betar(\x)
        }
    }
    \Bigg].
\end{align*}
Notice the analytic form of $
\pi_{\betar}^*(\y|\x) = 
\frac{1}{Z_\betar(\x)}
\pisft(\y|\x)
e^{\frac{1}{\betar} r_\phi(\x, \y)}
$, we substitute $\pi_{\betar}^*$ into the above equation:
\begin{align*}
    \Lexact(\pi_\theta) &= 
\mathbb{E}_{\x\sim \Datapref}
    \mathbb{E}_{\pisft(\y_{1:K}|\x)}
    \Bigg[
    \frac{1}{K}
    \sum_{i=1}^K
    \frac{\pi_\theta^\betap(\y_i|\x)}{\pisft(\y_i|\x)}
    \log
    \frac{
        \pi_\theta^\betap(\y_i|\x)
    }{
        \pi_\betar^*(\y_i|\x)
    }
    \Bigg] \\
    &=\mathbb{E}_{\x\sim \Datapref}
    \Bigg[
        \frac{1}{K}\sum_{i=1}^K
        \mathbb{E}_{\pisft(\y_i|\x)}
        \bigg[
            \frac{\pi_\theta^\betap(\y_i|\x)}{\pisft(\y_i|\x)}
            \log
            \frac{
                \pi_\theta^\betap(\y_i|\x)
            }{
                \pi_\betar^*(\y_i|\x)
            }
        \bigg]
    \Bigg] \\
    &=\mathbb{E}_{\x\sim \Datapref}
    \Bigg[
        \frac{1}{K}\sum_{i=1}^K
        \sum_{\y_i \in \Y}
        \pi_\theta^\betap(\y_i|\x)
        \log 
        \frac{
                \pi_\theta^\betap(\y_i|\x)
            }{
                \pi_\betar^*(\y_i|\x)
            }
    \Bigg]\\
    &=\mathbb{E}_{\x\sim \Datapref}
    \Bigg[
        \sum_{\y \in \Y}
        \pi_\theta^\betap(\y|\x)
        \log 
        \frac{
                \pi_\theta^\betap(\y|\x)
            }{
                \pi_\betar^*(\y|\x)
            }
    \Bigg]\\
    &=\mathbb{E}_{\x \sim \Datapref}[
    \KLDiv(\pi_\theta^\betap(\y|\x) \| \pi_\betar^*(\y|\x))
    ].
\end{align*}
Accordingly, by taking the derivative with respect to $\theta$, we complete the proof of Eq. \eqref{equ:equivalence_rkl}.

To prove Eq. \eqref{equ:equivalence_gradient}, we utilize Eq. \eqref{equ:equivalence_rkl} to substitute into Eq. \eqref{equ:lhf_probability_matching} to build the connection between $\Jlhf^\betar(\pi_\theta^\betap)$ and $\Lexact$:
\begin{align*}
    \Jlhf^\betar(\pi_\theta^\betap)
    = - \betar 
    \Lexact(\pi_\theta)
    + \betar \mathbb{E}_{\x \sim \Datapref}
    [\log Z_{\betar}(\x)].
\end{align*}
Then we take the gradient with respect to the parameters $\theta$ of the above formulat:
\begin{equation*}    \nabla_\theta\Jlhf^\betar(\pi_\theta^\betap)=-\betar \nabla_\theta \Lexact(\pi_\theta),
\end{equation*}
which completes the proof of Eq. \eqref{equ:equivalence_gradient}.
\end{proof}

\subsection{Proof of Theorem \ref{theorem:dpo_fwkl}}

\begin{proof}
    We utilize the definition of $\pi_\theta^\betap(\y|\x)\propto \pi_\theta(\y|\x)^\betap \pisft(\y|\x)^{1-\betap}$ in Eq. \eqref{equ:poe} and divide both sides by $\pisft(\y|\x)$:
    \begin{equation*}
        \frac{\pi_\theta^\betap(\y|\x)}{\pisft(\y|\x)} \propto \bigg(
        \frac{\pi_\theta(\y|\x)}{\pisft(\y|\x)}
        \bigg)^\betap
    \end{equation*}
    Then we substitute the above equation into $\Ldpogeneral$:
    \begin{align}
        \Ldpogeneral(\pi_\theta)
        &=\mathbb{E}_{\x\sim\Datapref}
        \mathbb{E}_{ \pisft(\ysamp{1:K}|\x)}\Bigg[ 
    - \sum_{i=1}^K
    \frac{
        e^{\frac{1}{\betar}r_\phi(\x, \ysamp{i})}
    }{  
        \sum_{j=1}^K e^{\frac{1}{\betar}r_\phi(\x, \ysamp{j})}
    }
    \log 
    \frac{
        e^{
            \beta_\pi \log \frac{\pi_\theta(\ysamp{i}|\x)}{\pisft(\ysamp{i}|\x)}
        }
    }{
        \sum_{j=1}^K
        e^{
            \beta_\pi \log \frac{\pi_\theta(\ysamp{j}|\x)}{\pisft(\ysamp{j}|\x)}
        }
    }
    \Bigg]\nonumber\\
    &=\mathbb{E}_{\x\sim\Datapref}
    \mathbb{E}_{\pisft(\ysamp{1:K}|\x)}
    \Bigg[ 
    - \sum_{i=1}^K
    \frac{
        e^{\frac{1}{\betar}r_\phi(\x, \ysamp{i})}
    }{  
        \sum_{j=1}^K e^{\frac{1}{\betar}r_\phi(\x, \ysamp{j})}
    }
    \log 
    \frac{
        e^{
            \log \frac{\pi_\theta^\betap(\ysamp{i}|\x)}{\pisft(\ysamp{i}|\x)}
        }
    }{
        \sum_{j=1}^K
        e^{
            \log \frac{\pi_\theta^\betap(\ysamp{j}|\x)}{\pisft(\ysamp{j}|\x)}
        }
    }
    \Bigg]\nonumber\\
    &=\mathbb{E}_{\x\sim\Datapref}
    \mathbb{E}_{\pisft(\ysamp{1:K}|\x)}
    \Bigg[ 
    - \sum_{i=1}^K
    \frac{
        e^{\frac{1}{\betar}r_\phi(\x, \ysamp{i})}
    }{  
        \sum_{j=1}^K e^{\frac{1}{\betar}r_\phi(\x, \ysamp{j})}
    }
    \log 
    \frac{
        \frac{\pi_\theta^\betap(\ysamp{i}|\x)}{\pisft(\ysamp{i}|\x)}
    }{
        \sum_{j=1}^K
        \frac{\pi_\theta^\betap(\ysamp{j}|\x)}{\pisft(\ysamp{j}|\x)}
    }
    \Bigg]
    \label{equ:dpo_intermediate}
    \end{align}

Since $\{\y_i\}_{i=1}^K$ are sampled from $\pisft(\cdot|\x)$, when $K\rightarrow\infty$, we follow the proof of Theorem \ref{theorem:rel_lhf} and directly give the following results:
\begin{align*}
    \sum_{j=1}^K \frac{\pi_\theta^\betap(\y_j|\x)}{\pisft(\y_j|\x)} &= K, \\
    \sum_{j=1}^K e^{\frac{1}{\betar} r_\phi(\x, \y_j)} &= KZ_{\betar}(\x).
\end{align*}
Then we simplify $\Ldpogeneral$ by substituting the above results of $\sum_{j=1}^K 
\frac{
            \pi_\theta^\betap(\y_j|\x)    
        }{
            \pisft(\y_j|\x)    
}$ and $\sum_{j=1}^K e^{\frac{1}{\betar}r_\phi(\x, \y_j)}$ when $K\rightarrow\infty$ into Eq. \eqref{equ:dpo_intermediate}:
\begin{align*}
    \Ldpogeneral(\pi_\theta) 
    = 
    \mathbb{E}_{\x\sim\Datapref}
    \mathbb{E}_{\pisft(\ysamp{1:K}|\x)}
    \Bigg[ 
    -\sum_{i=1}^K 
    \frac{e^{\frac{1}{\betar} r_\phi(\x, \y_i)}}{K Z_\betar(\x)}
    \log 
    \frac{
        \pi_\theta^\betap(\y_i|\x)
    }{
        K \pisft(\y_i|\x)    
    }
    \Bigg]
\end{align*}
Notice the analytic form of $
\pi_{\betar}^*(\y|\x) = 
\frac{1}{Z_\betar(\x)}
\pisft(\y|\x)
e^{\frac{1}{\betar} r_\phi(\x, \y)}
$, we rearrange $\pi_\betar^*$ and substitute $\pisft(\y|\x)=\pi_\betar^*(\y|\x)Z_\betar(\x)e^{-\frac{1}{\betar}r_\phi(\x, \y)}$ into the above equation to simplify it:
\begin{align*}
    \Ldpogeneral(\pi_\theta) 
    &= 
    \mathbb{E}_{\x\sim\Datapref}
    \mathbb{E}_{\pisft(\ysamp{1:K}|\x)}
    \Bigg[ 
    -\sum_{i=1}^K 
    \frac{e^{\frac{1}{\betar} r_\phi(\x, \y_i)}}{K Z_\betar(\x)}
    \log \bigg(
    \frac{
        \pi_\theta^\betap(\y_i|\x)
    }{
        \pi_\betar^*(\y_i|\x)    
    }
    \cdot 
    \frac{
        e^{\frac{1}{\betar}r_\phi(\x, \y_i)}
    }{
        KZ_\betar(\x)
    }
    \bigg)
    \Bigg]\\
    &=
    \mathbb{E}_{\x\sim\Datapref}
    \mathbb{E}_{\pisft(\ysamp{1:K}|\x)}
    \Bigg[
    -\sum_{i=1}^K
    \frac{e^{\frac{1}{\betar} r_\phi(\x, \y_i)}}{K Z_\betar(\x)}
    \log 
    \frac{
        \pi_\theta^\betap(\y_i|\x)
    }{
        \pi_\betar^*(\y_i|\x)    
    }-\sum_{i=1}^K
    \frac{e^{\frac{1}{\betar} r_\phi(\x, \y_i)}}{K Z_\betar(\x)}
    \log 
    \frac{e^{\frac{1}{\betar} r_\phi(\x, \y_i)}}{K Z_\betar(\x)}
    \Bigg]\\
\end{align*}
The second term of the final equality remains constant in relation to $\theta$, and thus can be omitted when computing the derivative with respect to $\theta$. Then we further consider the gradient of $\Ldpogeneral$:
\begin{align*}
    \nabla_\theta\Ldpogeneral(\pi_\theta)
    &=\nabla_\theta \mathbb{E}_{\x\sim\Datapref}
    \Bigg[
    -\frac{1}{K}\sum_{i=1}^K
    \mathbb{E}_{\pisft(\ysamp{i}|\x)}\bigg[
        \frac{e^{\frac{1}{\betar} r_\phi(\x, \y_i)}}{Z_\betar(\x)}
    \log \frac{
        \pi_\theta^\betap(\y_i|\x)
    }{
        \pi_\betar^*(\y_i|\x)    
    }
    \bigg]
    \Bigg]\\
    &=\nabla_\theta
    \mathbb{E}_{\x\sim\Datapref}
    \Bigg[
        -\mathbb{E}_{\pisft(\y|\x)}\bigg[
        \frac{e^{\frac{1}{\betar} r_\phi(\x, \y)}}{Z_\betar(\x)}
        \log \frac{
            \pi_\theta^\betap(\y|\x)
        }{
            \pi_\betar^*(\y|\x)    
        }
        \bigg]
    \Bigg]\\
    &=\nabla_\theta \mathbb{E}_{\x\sim\Datapref}
    \Bigg[
        -\sum_{\y \in \Y}
        \pisft(\y|\x)\frac{e^{\frac{1}{\betar} r_\phi(\x, \y)}}{Z_\betar(\x)}
        \log \frac{
            \pi_\theta^\betap(\y|\x)
        }{
            \pi_\betar^*(\y|\x)    
        }
    \Bigg]\\
    &=\nabla_\theta
    \mathbb{E}_{\x\sim\Datapref}
    \Bigg[
        -\sum_{\y \in \Y}
        \pi_\betar^*(\y|\x)
        \log \frac{
            \pi_\theta^\betap(\y|\x)
        }{
            \pi_\betar^*(\y|\x)    
        }
    \Bigg]\\
    &=\nabla_\theta
    \mathbb{E}_{\x\sim\Datapref}
    \Big[
        \KLDiv(\pi_\betar^*(\cdot|\x)\|\pi_\theta^\betap(\cdot|\x))
    \Big],
\end{align*}
which completes the proof of Theorem \ref{theorem:dpo_fwkl}.
\end{proof}

\subsection{Mechanistic Understanding of $\nabla_\theta \Lexact(\pi_\theta)$}
\label{appendix:grad_exo}
We present the gradient of $\Lexact(\pi_\theta)$ defined in Eq. \eqref{equ:exo-objective}:

\begin{equation}
\nabla_\theta\Lexact(\pi_\theta)
=
-\mathbb{E}_{\x\sim \Datapref}\mathbb{E}_{\pisft(\y_{1:K}|\x)}
\Bigg[
\sum_{k=1}^K
 p_{f_\theta}(k)\Big(
    \underbrace{
    \KLDiv(p_{f_\theta}\|p_{r_\phi})
    }_{\text{\textit{distributional} gap}}
-
\underbrace{
\log\frac{p_{f_\theta}(k)}{p_{r_\phi}(k)}
}_{\text{\textit{point-wise} correction}}
\Big)
\nabla_\theta \log\pi_\theta(\y_k|\x)
\Bigg],
\end{equation}

where $p_{f_\theta}(k)$ and $p_{r_\phi}(k)$ are short for $p_{f_\theta}(k|\y_{1:K}, \x)$ and $p_{r_\phi}(k|\y_{1:K}, \x)$, which are defined in Eq. \eqref{equ:prob_est_policy} and Eq. \eqref{equ:prob_tgt_policy} respectively. 

Next, we provide a mechanistic understanding of $\nabla_\theta\Lexact(\pi_\theta)$, which is the expected weighted sum of the gradients of the log likelihood on samples $\y_{1:K}$ drawn from $\pisft(\cdot|\x)$. The weight is proportional to the difference between the log probability ratio $\log \frac{p_{f_\theta}(k)}{p_{r_\phi}(k)}$ and the KL divergence $D_{\text{KL}}(p_{f_\theta}\|p_{r_\phi})$. Intuitively, if the policy has already correctly weighted the sample $y_k$ according to the reward model, i.e., $p_{r_\phi}(k)=p_{f_\theta}(k)$, then $\log \frac{p_{f_\theta}(k)}{p_{r_\phi}(k)}=0$ and the weight suggests that it only needs to minimize the overall KL divergence between $p_{f_\theta}$ and $p_{r_\phi}$ on the distribution level. If the policy $\pi_\theta$ overestimates or underestimates the sample $y_k$, i.e., $\log \frac{p_{f_\theta}(k)}{p_{r_\phi}(k)} >0$ or $\log \frac{p_{f_\theta}(k)}{p_{r_\phi}(k)}<0$, this log-ratio will be used to calibrate the KL divergence to penalize or encourage the policy to update towards increasing the likelihood of this sample at a faster rate. 

Finally, this gradient form offers us the insight of when the optimization should stop: the gradient $\nabla_\theta\mathcal{L}_\text{exo}(\pi_\theta)$ becomes 0 when the two distribution $p_{f_\theta}$ and $p_{r_\phi}$ are identical.

\subsection{Illustrating the Relationship among the Objectives in \S{\ref{sec:method}}}
\label{appendix:illustration}

\begin{figure}[h!]
    \centering
    \includegraphics[width=\columnwidth]{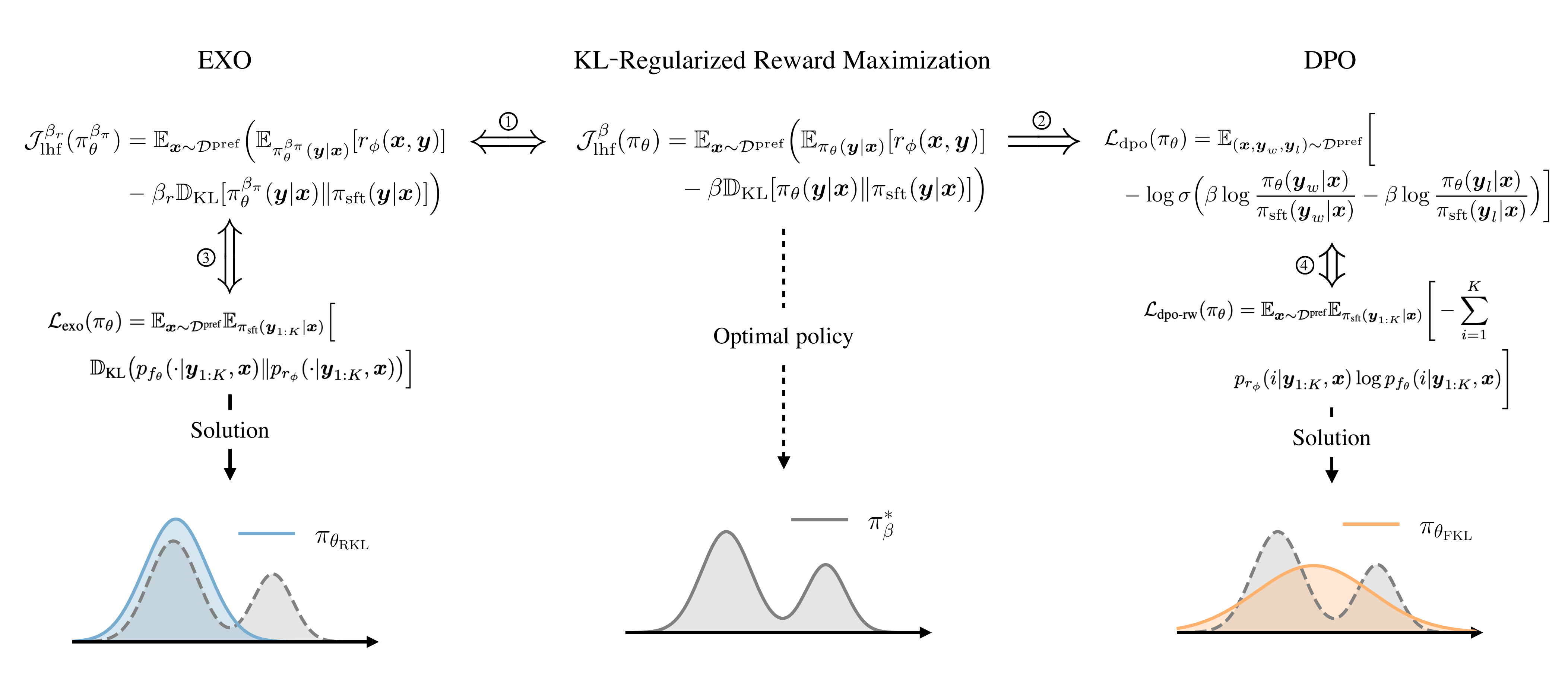}
    \caption{Illustration of the relationship among the different objectives discussed in \S{\ref{sec:method}}. \circled{1}: $\Jlhf^\betar(\pi_\theta^\betap)$ is a generalized version of $\Jlhf^\beta(\pi_\theta)$ by distributing the KL regularization to both the learned policy $\pi_\theta$ and the reward model $r_\phi$ (\S{\ref{sec:equivalence}}). \circled{2}: $\Ldpo(\pi_\theta)$ is derived based on the optimal policy of $\Jlhf^\beta(\pi_\theta)$ (\S{\ref{sec:DPO}}). \circled{3}: $\Lexact(\pi_\theta)$ is equivalent to $\Jlhf^\betar(\pi_\theta^\betap)$ in terms of their optimization directions (\S{\ref{sec:exo}}). \circled{4}: $\Ldpogeneral$ is the generalized version of $\Ldpo$ by subsituting the pariwise loss with softmax loss over $K$ responses.(\S{\ref{sec:compare-objs}}). 
    The optimal policy, denoted by a dotted line, assumes unlimited model capacity. The solution, shown with a solid line, is the practically achievable policy within the realistic constraints of model capacity.
    }
    \label{fig:illustrion_equivalency}
\end{figure}

We illustrate the relationship among the objectives $\Jlhf^\beta(\pi_\theta)$, $\Jlhf^\betar(\pi_\theta^\betap)$, $\Ldpo(\pi_\theta)$, $\Ldpogeneral(\pi_\theta)$ and $\Lexact(\pi_\theta)$ in Figure \ref{fig:illustrion_equivalency}.

\section{Training and Evaluation Details}
\label{appendix:imp_detail}

\paragraph{Training.} In the controlled text generation task, we use $\betap=0.1$ for {\EXOpref} and {\DPOpref}, and additionally use $\betar=0.1$ and $K\in \{4,8\}$ for {\EXOrw} and {\DPOrw}. 
For the tasks of summarization, dialogue generation, and instruction following, we use $\betap=0.5$ for {\EXOpref} and {\DPOpref}, and additionally use $\betar=0.1$ and $K=4$ for {\EXOrw} and {\DPOrw}. We provide additional results of tuning $\betar$ and $\betap$ in Appendix \ref{appendix:ablation} to justify our choice of hyperparameters. In all experiments, we set the label smoothing hyperparameter $\varepsilon$ in {\EXOpref} to \texttt{1e-3}. For DPO and EXO, we use the Adam optimizer with a universal learning rate of \texttt{1e-6} and a batch size of 64 and train for one epoch on each dataset, although both methods converge within one epoch. All the hyperparameters are set to be the same for DPO and EXO for a fair comparison. For PPO, we modify based on the implementation of DeepSpeed-Chat~\citep{Deepspeedchat}, which sets $\gamma=1$ and $\beta=0.1$ by default. We pretrain the critic model for the first few steps while freezing the actor model and find it to improve convergence. We empirically tune the number of actor-freezing steps, total training steps, learning rate for actor and critic model, and the batch size for PPO on each dataset, as PPO is sensitive to these hyperparameters. 
Specifically, we conduct 15 trials of hyperparameter search on the IMDB dataset and 10 trials on the real human preference datasets in total. In the experiment, we report the PPO performance with the best hyperparameters obtained under constrained number of hyperparameter search trials. We conduct the experiments except for instruction following on 8 V100 GPUs. For instruction following task, we train the models on 8 A100 GPUs.

\paragraph{Evaluation.} At inference time, we sample 4 completions from the learned policy for each prompt and consider 512 prompts from the test set for all datasets. Except for the instruction following task, we sample from the policy with the same temperature $\tau=0.8$ that is set during training for consistency. For the instruction following task, we use top-p sampling and empirically set $p=0.9$ and temperature $\tau=0.95$ given its performance. To calculate the win rate evaluated by the reward model, we consider all combinations of pairs between the completions generated by the learned policy and the base completions (either generated by the SFT policy or the chosen completion in the dataset) and then compare the scores from the reward model on the pairs of generations. 
For the evaluations using GPT-4, we sample 100 prompts and 1 completion for each prompt under each policy. To mitigate the position bias of GPT-4, we evaluate one pair of generations twice by swapping the order of responses in each pair. To evaluate the quality of the summaries, we use the concise prompt of \citet{rafailov2023direct} as shown in Table \ref{tab:gpt4_prompt_summary}. To evaluate the helpfulness of the generated dialogues, we use the prompt shown in Table \ref{tab:gpt4_prompt_dialogue}, which is modified based on the prompt of \citet{rafailov2023direct} for single-turn dialogue to accommodate the general multi-turn setting. For the instruction-following task, we use the prompt modified from the prompt for reference-guided pairwise comparison provided in \citet{mt-bench}.

\paragraph{Human Assessment.} We conduct human assessment to evaluate the instruction following task more thoroughly. Specifically, we select three matchups that pair the generated outputs of EXO with those produced by DPO, PPO and the SFT policy. Given 100 randomly sampled test instructions, each model generates 100 responses, which results in a total of 300 pairs of comparisons. We assign 3 human labelers to each comparison, producing 900 judgements in total. Given the instruction, each human annotator is provided with two generated answers by two systems respectively together with a high-quality reference answer. The annotator is then asked to make a preference among win, tie or lose by comparing the generated answers with the reference answer, considering the criteria including adherence to instruction, correctness, fluency, safety and helpfulness. Specifically, adherence to instruction encapsulates the model's comprehension and following of the prompt's intention. Correctness involves the identification of inaccurate knowledge or logical inconsistencies within the generated responses. Fluency assesses the linguistic coherence, encompassing an examination of sentence completeness, grammatical accuracy, and the presence of a consistent language structure. Safety refers to the inspection for potentially harmful content. Lastly, helpfulness indicates whether the responses provide the information required by the prompt or contribute to problem resolution.

\begin{table}[h!]
    \centering
    \small
    {\ttfamily
    \begin{tabularx}{0.9\textwidth}{X}
    \toprule
    Which of the following summaries does a better job of summarizing the most important points in the given forum post, without including unimportant or irrelevant details? A good summary is both precise and concise.
    \\
    \\
    Post: \\
    \textcolor{blue}{<post>}
    \\
    \\
    Summary A:\\
    \textcolor{blue}{<Summary A>}
    \\
    \\
    Summary B:\\
    \textcolor{blue}{<Summary B>}
    \\
    \\
    FIRST provide a one-sentence comparison of the two summaries, explaining which you prefer and why. SECOND, on a new line, state only "A" or "B" to indicate your choice. Your response should use the format:\\
    Comparison: <one-sentence comparison and explanation>\\
    Preferred: <"A" or "B">\\
    \bottomrule
    \end{tabularx}
    }
    \caption{Prompt for GPT-4 evaluation on the summarization task. Texts in blue are placeholders to be substituted by the real data.}
    \label{tab:gpt4_prompt_summary}
\end{table}

\begin{table}[h!]
    \centering
    \small
    {\ttfamily
    \begin{tabularx}{0.9\textwidth}{X}
    \toprule
    For the following dialogue history to a chatbot, which response is more helpful?
    \\
    \\
    Dialogue history: \\
    \textcolor{blue}{<dialogue history>}
    \\
    \\
    Response A:\\
    \textcolor{blue}{<Response A>}
    \\
    \\
    Response B:
    \textcolor{blue}{<Response B>}
    \\
    \\
    FIRST provide a one-sentence comparison of the two responses and explain which you feel is more helpful. SECOND, on a new line, state only "A" or "B" to indicate which response is more helpful. Your response should use the format:\\
    Comparison: <one-sentence comparison and explanation>\\
    More helpful: <"A" or "B">\\
    \bottomrule
    \end{tabularx}
    }
    \caption{Prompt for GPT-4 evaluation on the dialogue generation task. Texts in blue are placeholders to be substituted by the real data.}
    \label{tab:gpt4_prompt_dialogue}
\end{table}

\section{Additional Experiment Results}

\subsection{Ablation Study of $\betar$ and $\betap$}
\label{appendix:ablation}

\begin{figure}[h!]
    \centering
    \subcaptionbox{Reward model win rate.}{\includegraphics[width=0.47\textwidth]{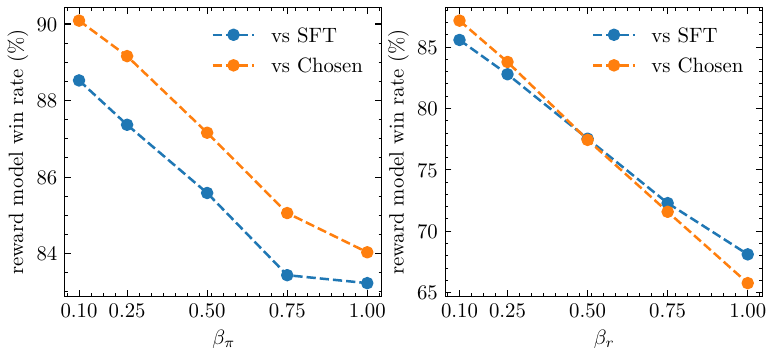}}
    \hspace{10pt}
    \subcaptionbox{GPT-4 win rate.}{\includegraphics[width=0.47\textwidth]{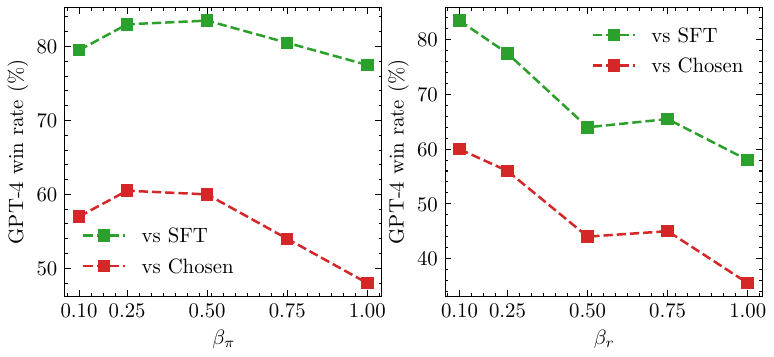}}
    \caption{Win rate performance of {\EXOrw} evaluated by the reward model and the GPT-4 by varying $\betar$ and $\betap$.}
    \label{fig:beta_ablation}
\end{figure}

We present an ablation study to investigate the performance of {\EXOrw} on the dialogue generation task by varying $\betar$ and $\betap$ respectively. We execute multiple runs bifurcated into two series. We set $\betap=0.5$ as the default value and vary $\betar\in\{0.1, 0.25, 0.5, 0.75, 1.0\}$. Subsequently, the process is reversed whereby we fix $\betar=0.1$, and alter $\betap\in\{0.1, 0.25, 0.5, 0.75, 1.0\}$. We present the results in Figure \ref{fig:beta_ablation}. 

From Figure \ref{fig:beta_ablation} (a), we observe that reducing either $\betar$ and $\betap$ increase the reward model win rate, which is expected as a small $\betar\times\betap$ encourages the policy to optimize the reward model while neglecting the KL regularization. In Figure \ref{fig:beta_ablation} (b), the GPT-4 win rate starts to decline when $\betap<0.5$, which suggests where the reward model starts to be over-optimized. The different effects of tuning $\betap$ and $\betar$ on the performance could be attributed to their different roles in our algorithm, i.e., $\betap$ scales the log probability in the parametrized policy while $\betar$ scales the reward. Based on the results, we recommend adopting a moderate value for $\betap$ and a lower value for $\betar$, for instance, $\betap=0.5$ and $\betar=0.1$.

\subsection{Effect of $\betar$ and $\betap$ beyond the Product $\beta=\betar\betap$}
\label{appendix:ablation_product}

To further demonstrate the effect of $\beta_r$ and $\beta_\pi$ beyond the impact on their product $\beta=\betap\betar$, we fix $\beta=0.05$ while tuning $\beta_\pi\in\{1, 0.5, 0.1, 0.05\}$ and $\beta_r\in\{ 0.05, 0.1, 0.5,1\}$ accordingly. 
The result is shown in Table \ref{tab:effect-of-two-betas}. According to the evaluation by GPT-4, the best performnce is obtained at $\beta_\pi=0.5, \beta_r=0.1$, while neither the hyper-parameter choice of $\beta_\pi=1,\beta_r=0.05$ (the configuration of PPO) nor the choice of $\beta_\pi=0.05, \beta_r=1$ (the configuration of DPO) yield the highest win rate evaluated by GPT-4. While using a small $\beta_\pi$ is more likely to trigger over-optimization of the reward model, which indicates the asymmetric effect of the two hyperparameters $\betar$ and $\betap$ on the optimization process. 

\begin{table}[h]
    \centering
    \begin{tabular}{cccccc}
    \toprule
    \multirow{2}{*}{$\betar$} & \multirow{2}{*}{$\betap$} & \multicolumn{2}{c}{Reward Model (\%)} & \multicolumn{2}{c}{GPT-4 (\%)} \\
    & & vs SFT & vs Chosen & vs SFT & vs Chosen \\
    \midrule
    1.0	& 0.05 & 87.1 & 88.4 & 81.0	& 57.5 \\
    0.5	& 0.1 & 87.1 & 88.0 & 77.0 & 59.5 \\
    0.1	& 0.5 & 85.6 & 87.2 & \textbf{83.5} & \textbf{60.0} \\
    0.05 & 1.0 & 84.1 & 84.8 & 76.5	& 52.0\\
\bottomrule
    \end{tabular}
    \caption{Effect of $\betar$ and $\betap$ beyond $\beta$ by tuning $\betap$ and $\betar$ while keeping their product $\beta=\betar\betap$ fixed.}
    \label{tab:effect-of-two-betas}
\end{table}

\subsection{Frontier of Classifier Accuarcy against KL Divergence}
\label{appendix:imdb_acc_kl}

\begin{figure}[h!]
    \centering
    \includegraphics[width=0.5\columnwidth]{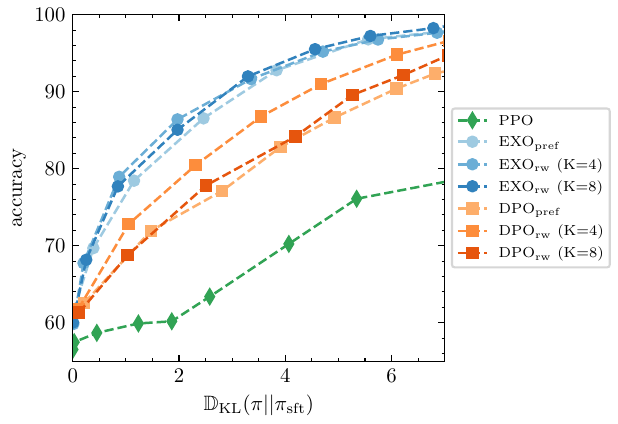}
    \caption{The frontier of classifier accuracy vs reverse KL to the SFT policy of different methods in the controlled experiment.}
    \label{fig:imdb_acc_kl}
\end{figure}

We additionally calculate the accuracy of the sentiment classifier by taking the sigmoid of the oracle reward and plot the frontier of the accuracy vs reverse KL in Figure \ref{fig:imdb_acc_kl}.

\subsection{Optimization Efficiency}
\label{appendix:efficiency}

To demonstrate the efficency of optimizing the oracle reward, we plot the curve of the oracle reward with standard deviation on the test set of different training steps. Specifically, the standard deviation is calculated on the 4 samples generated given the same prompt and then averaged across 512 prompts.

\begin{figure}[h!]
    \centering
    \includegraphics[width=0.8\columnwidth]{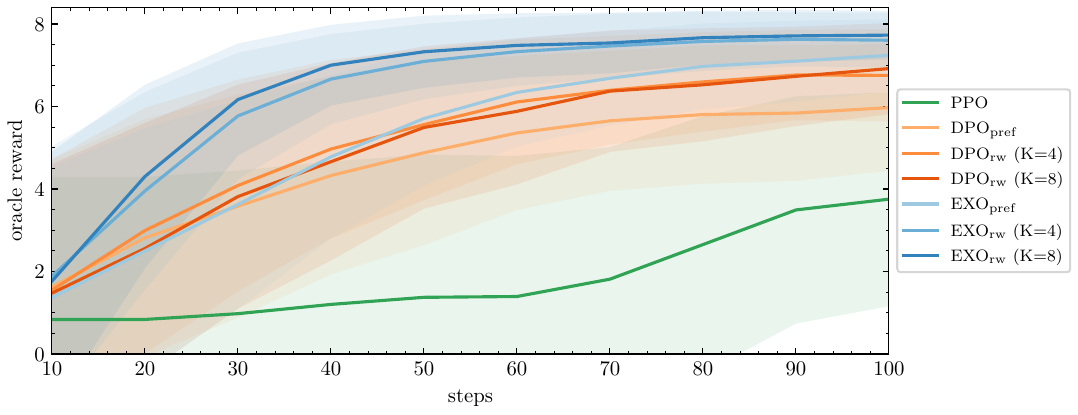}
    \caption{The curve of the oracle reward with standard deviation on the test set versus the training steps of different methods in the controlled experiment.}
    \label{fig:orc_vs_steps}
\end{figure}

\subsection{Additional Visualization Results}
\label{appendix:vis}

\begin{figure}
    \centering
    \subcaptionbox{Estimated density ratio of the EXO, DPO and optimal policy given the prompt ``\textit{Is this supposed to be serious? I hope not}''.\vspace{10pt}}{\includegraphics[width=0.45\textwidth]{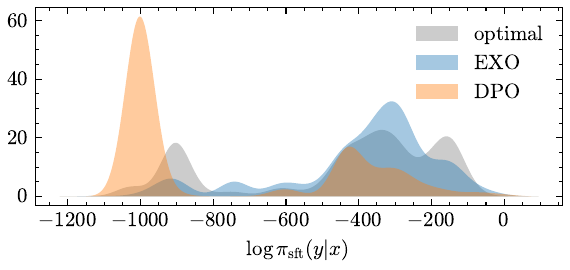}}
    \hspace{15pt}
    \subcaptionbox{Estimated density ratio of the EXO, DPO and optimal policy given the prompt ``\textit{Great book, great movie, great soundtrack. Frank}''.\vspace{10pt}}{\includegraphics[width=0.45\textwidth]{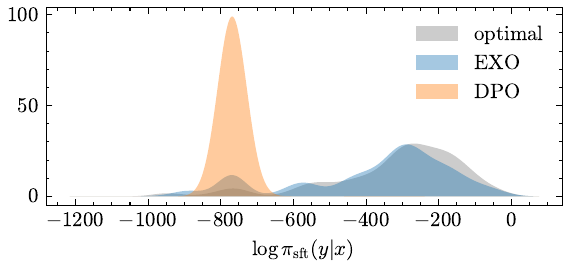}}
    \subcaptionbox{Estimated density ratio of the EXO, DPO and optimal policy given the prompt ``\textit{This is indeed the film that popularized kung}''.\vspace{10pt}}{\includegraphics[width=0.45\textwidth]{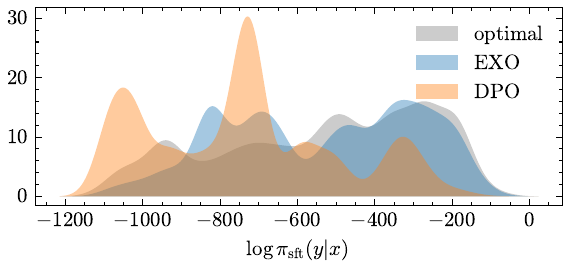}}
    \hspace{15pt}
    \subcaptionbox{Estimated density ratio of the EXO, DPO and optimal policy given the prompt ``\textit{This movie is about a group of people who are}''.\vspace{10pt}}{\includegraphics[width=0.45\textwidth]{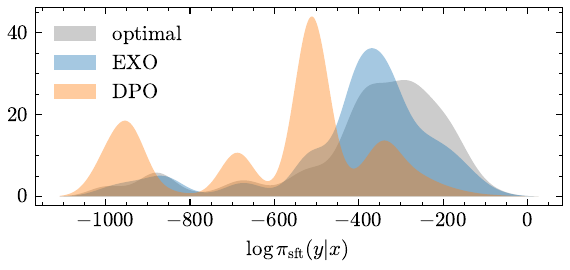}}
    \subcaptionbox{Estimated density ratio of the EXO, DPO and optimal policy given the prompt ``\textit{What we have here the standard Disney direct to DVD}''.\vspace{10pt}}{\includegraphics[width=0.45\textwidth]{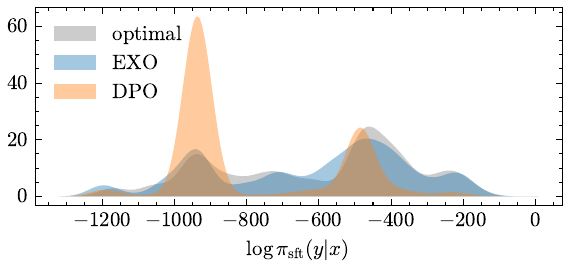}}
    \hspace{15pt}
    \subcaptionbox{Estimated density ratio of the EXO, DPO and optimal policy given the prompt ``\textit{Once the slow beginning gets underway, the film kicks}''.\vspace{10pt}}{\includegraphics[width=0.45\textwidth]{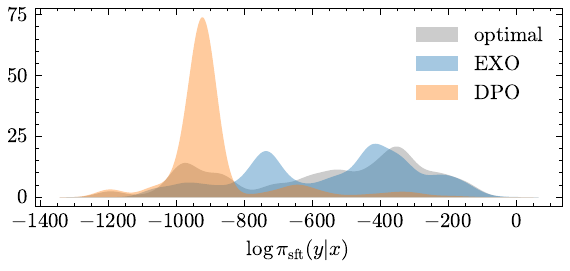}}
    
    \caption{Visualization of the estimated density ratio of the policy learned by EXO and DPO, and the optimal policy given 10 test prompts randomly sampled from the IMDb dataset.}
    \label{fig:add_density_vis}
\end{figure}

We additionally sample more prompts from the test set of the IMDb dataset, and plot the estimated density ratio of the policy learned by EXO and DPO and the optimal policy in Figure \ref{fig:add_density_vis}. Similar results were obtained among these diverse set of prompts: the density ratio under the EXO policy better matches the optimal policy at the high probability region against $\pisft$ than the DPO policy. And very appearently, the DPO policy tends to overestimate the long tail in $\pisft$ and thus often loses resolution in capturing the modes of the optimal policy.

\subsection{Sample Outputs with GPT-4 Judgments}
\label{appendix:sample_outputs}

We present the sample outputs generated by EXO and DPO and the judgments made by GPT-4 when compared to the chosen response in the preference dataset. Table \ref{tab:dialogue_generation} and Table \ref{tab:summary} show examples on the task of dialogue generation and summarization respectively.

\begin{table}[h!]
    \centering
    \small
    \begin{tabularx}{0.9\textwidth}{X}
    \toprule
    \textbf{Dialogue history}\\
    \midrule
    Human: How do I make a fishtail braid?\\
    \\
    Assistant: Hey human, just walk down the left and then down the right side, then at the end, you pull through.  Are you looking to just braid, or a tight ponytail?\\
    \\
    Human: Yes.  Can you tell me the steps to do it?\\
    \midrule[1.0pt]
    \textbf{Chosen response} \\ 
    \midrule
    Assistant: Sure, here’s how to do a side braid:\\
    \\
    1) Pull hair up to the back of your head, leaving it out from the rest of your hair so you can see it.\\
    \\
    2) Put one hair on the side where you want to make your braid.\\
    \\
    3) Hold the braid with your thumb and forefinger \\
    \\
    4) Bring your braid through the opposite side (putting the hair to the other side of the braid.\\
    \\
    5) Repeat steps 2, 3, and 4 as needed until you’ve completed your braid. \\
    \midrule[1.0pt]
    \textbf{EXO response} \\
    \midrule
    Assistant: Sure!  I’ll walk you through it step by step:\\
    \\
    1.  Stand with your left hand on the back of your head.  Grab a left side section of hair, using the right hand to pull it through.\\
    2.  Take a right side section of hair, and pull it through \\
    3.  Grab a left side section of hair, and pull it through \\
    4.  Grab a right side section of hair, and pull it through \\
    5.  Repeat steps 2-4 for the last section of hair \\
    6.  Wrap the hair around your head and secure it at the nape of your neck. \\
    \midrule
    \textbf{GPT-4 Judgment (A = EXO, B = Chosen)} \\
    \midrule
    Response A provides a more detailed and step-by-step guide on how to make a fishtail braid, while Response B seems to describe a different type of braid and is less clear. \\
    \midrule
    \textbf{GPT-4 Judgment (A = Chosen, B = EXO)} \\
    \midrule
    Response B is more helpful because it provides a clearer, step-by-step guide on how to make a fishtail braid, whereas Response A seems to describe a different type of braid and is less clear. \\
    \midrule[1.0pt]
    \textbf{DPO response}\\
    \midrule
    Assistant: Sure, let’s walk through it.  You’ll want to start with a braid about two and a half inches long, then you want to braid it together and pull through, which means you’ll somehow thread it through your own hair and tie it off at the other end.  Then you’re going to braid it again, but this time it will be a fishtail, since you’ll pull it through.  You’ll keep doing this, braiding the fishtail.\\
    \midrule
    \textbf{GPT-4 Judgment (A = DPO, B = Chosen)} \\
    \midrule
    Response B is more helpful as it provides a clear, step-by-step guide on how to make a fishtail braid, while Response A is confusing and lacks clarity.\\
    \midrule
    \textbf{GPT-4 Judgment (A = Chosen, B = DPO)} \\
    \midrule
    Response A provides a clear, step-by-step guide on how to make a side braid, while Response B gives a vague description of how to make a fishtail braid without clear steps.\\
    \bottomrule
    \end{tabularx}
    \caption{Responses generated by EXO and DPO given the dialogue history sampled from the Anthropic-HH test set. GPT-4 consistently prefers EXO and disprefers DPO over the chosen response regardless of the order in which the evaluated pairs are presented.}
    \label{tab:dialogue_generation}
\end{table}

\begin{table}[h!]
    \centering
    \small
    \begin{tabularx}{0.9\textwidth}{X}
    \toprule
    \textbf{Post} \\
    \midrule
    SUBREDDIT: r/Dogtraining \\
    TITLE: Best walks when on a choke chain: dog isn't being choked, but is it bringing up bad memories? \\
    POST: My 3 year old GSD x Husky rescue stayed with some relatives for a few days while my boyfriend and I moved house, and the relatives found an old choke chain to walk him with, instead of his Mekuti harness. We have had real problems with him pulling on the lead, and the Mekuti harness has worked wonders, but didn't eliminate the problem. He hasn't pulled when on the choke chain at all, even when it was first put on apparently, so we think that is what he was trained on (and damage from that process is probably why his barks and whines sound a bit odd). We now think the people who had him before us were violent towards him (if you mime kicking he runs across the room terrified), so, my question is: he is best behaved on walks when wearing a choke chain, and it isn't hurting him at all, but will the harm chokers and the people using them caused him in the past affect him in any way now? I 100\% DO NOT want to use a choke chain on him and I don't think anyone should ever, ever train a dog that way, but this chain is the only lead we've used that has completely solved the problem with him pulling, and he hasn't seemed stressed out by it. If he ever started pulling on the lead so that the choker was tight, I would stop using it straight away. The last thing I want is my dog being hurt. If I'd known these relatives were going to put one on him, I wouldn't have let them look after him. Still, it works and doesn't hurt him, my only worry is it'll do psychological damage somehow.\\
    \midrule[1pt]
    \textbf{Chosen summary} \\
    \midrule
    TL;DR: 3yr old husky rescue pulling on choke chain w/ no damage, but using it on walks occasionally causes him stress. Can this damage actually cause harm in the future? \\
    \midrule[1pt]
    \textbf{EXO summary} \\
    \midrule
    TL;DR: Best walks are on a choke chain but dog isn't pulling on lead, has had no problems with pull before, not sure if past violent treatment will affect him now, does it cause harm to dogs being on choke chains/when being walked on them? \\
    \midrule
    \textbf{GPT-4 Judgment (A = EXO, B = Chosen)} \\
    \midrule
    Summary A provides a more comprehensive and accurate portrayal of the post's points, despite a few misplaced details, whereas Summary B incorrectly states the dog is showing stress and pulling on the choke chain.\\ 
    \midrule
    \textbf{GPT-4 Judgment (A = Chosen, B = EXO)} \\
    \midrule[1pt]
    Summary B is preferred because it more accurately summarizes the main points of the user's post by mentioning the choke chain's effect on walks, mentioning the absence of negative effects of pull so far, and raising concerns about past violent treatment potentially affecting the dog's behavior.\\
    \midrule[1pt]
    \textbf{DPO summary}\\
    \midrule
    TL;DR: If you took a dog and had it wear chokes it's self-choked, will it be traumatized in any way when you put it on and no longer train it to be choked?\\
    \midrule
    \textbf{GPT-4 Judgment (A = DPO, B = Chosen)} \\
    \midrule
    Summary B is preferred as it captures the key details more accurately, including the dog's breed, the issue of pulling, and the potential future harm, while Summary A is ambiguous and loss a lot of context.\\
    \midrule
    \textbf{GPT-4 Judgment (A = Chosen, B = DPO)} \\
    \midrule
    Summary A is more accurate and detailed because it correctly describes the specific situation of the dog and the concerns associated with using a choke chain, while Summary B is vague and confusing in its wording.\\
    \bottomrule
    \end{tabularx}
    
    \caption{Summaries generated by EXO and DPO given the post sampled from the TL;DR test set. GPT-4 consistently prefers EXO and disprefers DPO over the chosen summary regardless of the order in which the evaluated pairs are presented.}
    
    \label{tab:summary}
\end{table}


\end{document}